\batchmode
\documentclass[twocolumn,journal]{IEEEtran}
\usepackage[T1]{fontenc}
\usepackage{array}
\usepackage{multirow}
\usepackage{graphicx}
\usepackage[unicode=true,
 bookmarks=true,bookmarksnumbered=true,bookmarksopen=true,bookmarksopenlevel=1,
 breaklinks=false,pdfborder={0 0 0},backref=false,colorlinks=false]
 {hyperref}
\hypersetup{pdftitle={Learning Effective RGB-D Representations for Scene Recognition},
 pdfauthor={Xinhang Song, Shuqiang Jiang, Luis Herranz, Chengpeng Chen}}

\makeatletter

 \let\oldforeign@language\foreign@language
 \DeclareRobustCommand{\foreign@language}[1]{%
   \lowercase{\oldforeign@language{#1}}}

\usepackage[caption=false,font=footnotesize]{subfig}
\usepackage{tablefootnote}
\usepackage{threeparttable}

\usepackage{algorithmic}
\usepackage{algorithm}

\usepackage{tablefootnote}
\usepackage{threeparttable}

\makeatother

\begin{document}

\title{Learning Effective RGB-D Representations for Scene Recognition}

\author{Xinhang~Song,~Shuqiang Jiang{*} \textit{IEEE Senior Member},~Luis
Herranz, Chengpeng Chen\thanks{X. Song, S. Jiang C. Chen are with the Key Lab of Intelligent Information
Processing of Chinese Academy of Sciences (CAS), Institute of Computing
Technology, CAS, Beijing, 100190, China, are also with University
of Chinese Academy of Sciences, Beijing, 100049, China e-mail: xinhang.song@vipl.ict.ac.cn,
sqjiang@ict.ac.cn\protect \\
 L. Herranz is with the Computer Vision Center, Barcelona, Spain,
email:lherranz@cvc.uab.es.\protect \\
 C. Chen is with the Key Lab of Intelligent Information Processing
of Chinese Academy of Sciences (CAS), Institute of Computing Technology,
CAS, Beijing, 100190, China, and also with the University of Chinese
Academy of Sciences, Beijing, 100049, China e-mail: \{chengpeng.chen\}@vipl.ict.ac.cn.\protect \\
 {*}Corresponding author: Shuqiang Jiang. }}

\markboth{IEEE Transactions on Image Processing}{X. Song \MakeLowercase{\emph{et al.}}: Learning Effective RGB-D
Representations for Scene Recognition}

\IEEEpubid{}
\maketitle
\begin{abstract}
Deep convolutional networks (CNN) can achieve impressive results on
RGB scene recognition thanks to large datasets such as Places. In
contrast, RGB-D scene recognition is still underdeveloped in comparison,
due to two limitations of RGB-D data we address in this paper. The
first limitation is the lack of depth data for training deep learning
models. Rather than fine tuning or transferring RGB-specific features,
we address this limitation by proposing an architecture and a two-step
training approach that directly learns effective depth-specific features
using weak supervision via patches. The resulting RGB-D model also
benefits from more complementary multimodal features. Another limitation
is the short range of depth sensors (typically 0.5m to 5.5m), resulting
in depth images not capturing distant objects in the scenes that RGB
images can. We show that this limitation can be addressed by using
RGB-D videos, where more comprehensive depth information is accumulated
as the camera travels across the scenes. Focusing on this scenario,
we introduce the ISIA RGB-D video dataset to evaluate RGB-D scene
recognition with videos. Our video recognition architecture combines
convolutional and recurrent neural networks (RNNs) that are trained
in three steps with increasingly complex data to learn effective features
(i.e. patches, frames and sequences). Our approach obtains state-of-the-art
performances on RGB-D image (NYUD2 and SUN RGB-D) and video (ISIA
RGB-D) scene recognition.\end{abstract}


\IEEEpeerreviewmaketitle{}

\section{Introduction}

\IEEEPARstart{T}{he} goal of scene recognition is to predict scene
labels for visual data such as images and videos. Success in visual
recognition mainly depends on the features used to represent the input
data. Scene recognition in particular has benefited from recent developments
in data-driven representation learning, where massive image datasets
(ImageNet and Places \cite{Zhou2014}) provide the necessary amount
of data to effectively train complex convolutional neural networks
(CNNs) \cite{Krizhevsky2012,Simonyan2015} with millions of parameters.
The features extracted from models pretrained with those datasets
are generic and powerful enough to obtain state-of-the-art performance
in relevant scene benchmarks (e.g., MIT indoor 67 \cite{Quattoni2009},
SUN397 \cite{Xiao2010}), just using an SVM \cite{Jeff2014} or fine-tuning,
and outperforming earlier handcrafted paradigms (e.g., SIFT, HOG, bag-of-words).

In parallel, low cost depth sensors can capture depth information
that complements RGB data. Depth can provide valuable information
to model object boundaries and understand the global layout of the
scene. Thus, RGB-D models should improve recognition over mere RGB
models. However, RGB-D data needs to be captured with a specialized
and relatively complex setup \cite{Silberman2011,Song2015a} (in contrast
to RGB data that can be collected by crawling the web). For this reason,
RGB-D datasets are orders of magnitude smaller than the largest RGB
datasets, also with much fewer categories. Since depth images somewhat
resemble some aspects of RGB images (specially in certain color codings),
shapes and objects can be often identified in both RGB and depth images
(see Fig.~\ref{fig:motivation}). This motivates the common practice
of leveraging the architecture and parameters of a deep network pretrained
on large RGB datasets (e.g., ImageNet, Places) to then fine tune two
separate RGB and depth branches with the corresponding modality-specific
images from the target set. The two branches are then combined in
the final RGB-D model. This is the main approach used in recent works
\cite{Song2015a,Wang_2016_CVPR,Zhu_2016_CVPR,Gupta_2016_CVPR,wang2016correlated,eitel2015multimodal}.

However, relying on networks trained for RGB data to build depth features
seems to be an inherent limitation. The question is whether fine tuning
is the best possible solution given the limited depth data. Here we
challenge the usual assumption that learning depth features from scratch
with current limited data is still less effective. In fact, we show
that a significantly smaller network trained in a two-step process
with patches pretraining via weak supervison can effectively learn
more powerful depth features, more complementary to RGB ones, and
thus provide higher gains in RGB-D models. We show that this weakly-supervised
pretraining stage is critical to obtain powerful depth representations,
even more effective than those transferred from deeper RGB networks.

\begin{figure}[t]
\begin{centering}
\setlength{\tabcolsep}{1pt} \renewcommand{\arraystretch}{1}
\par\end{centering}

\begin{centering}
\begin{tabular}{ccc}
\includegraphics[width=0.35\columnwidth]{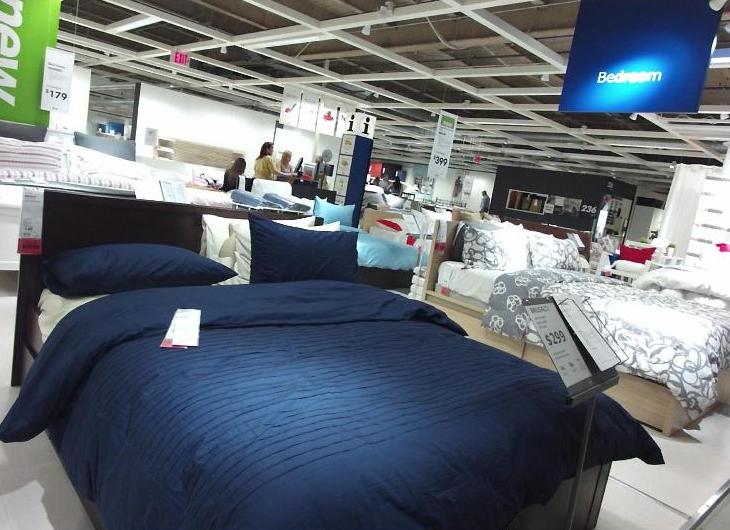} &
\includegraphics[width=0.35\columnwidth]{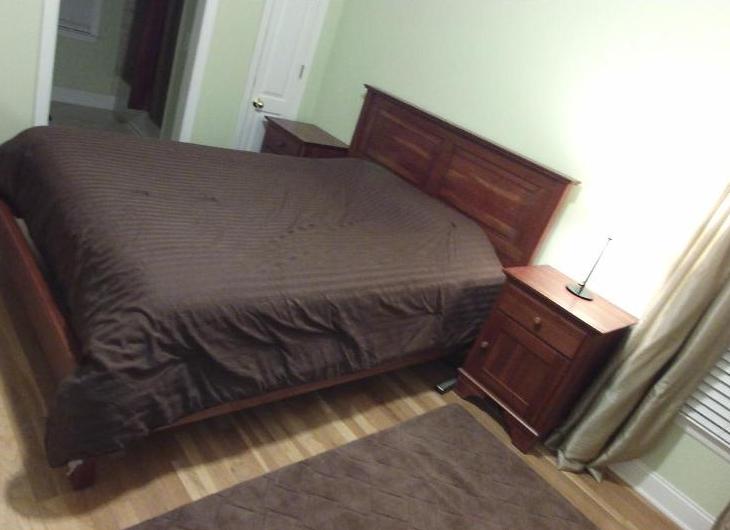} &
\multirow{1}{*}{RGB}\tabularnewline
\includegraphics[width=0.35\columnwidth]{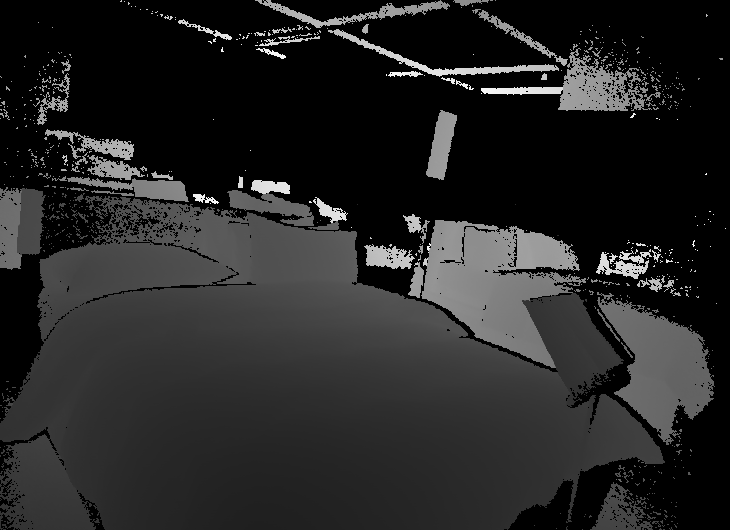} &
\includegraphics[width=0.35\columnwidth]{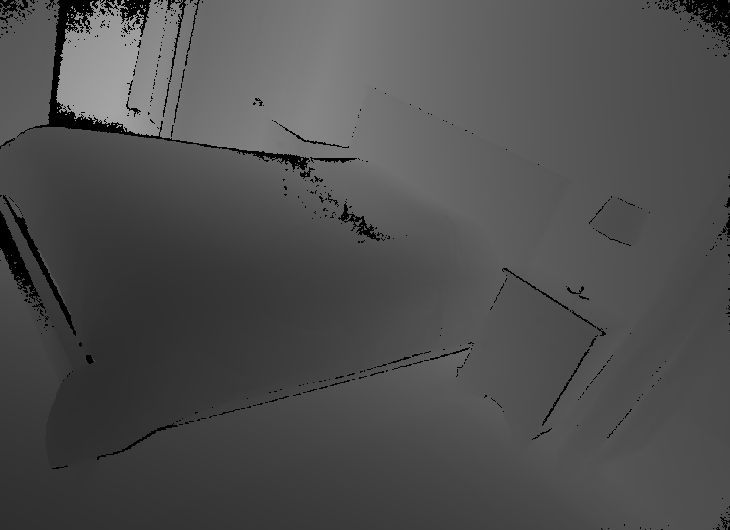} &
Depth\tabularnewline
(a) Furniture store &
(b) Bedroom &
\tabularnewline
\includegraphics[width=0.35\columnwidth]{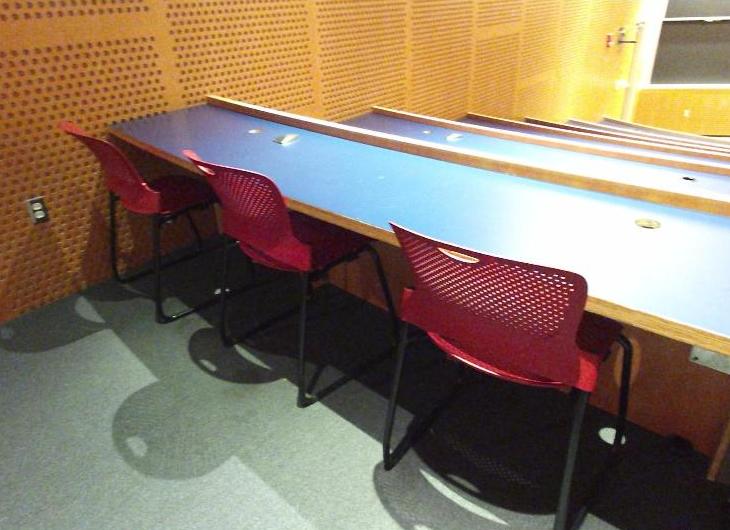} &
\includegraphics[width=0.35\columnwidth]{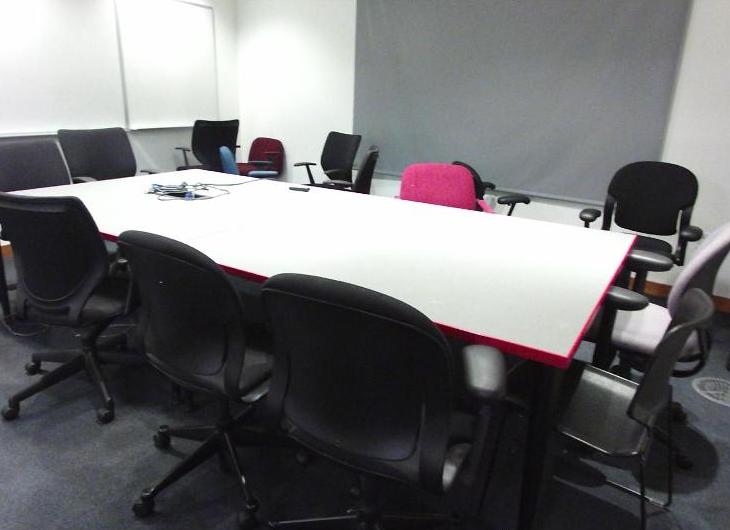} &
RGB\tabularnewline
\includegraphics[width=0.35\columnwidth]{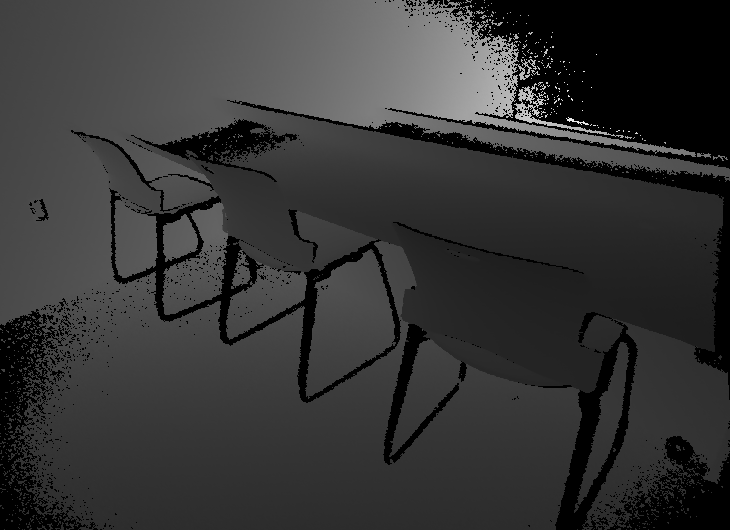} &
\includegraphics[width=0.35\columnwidth]{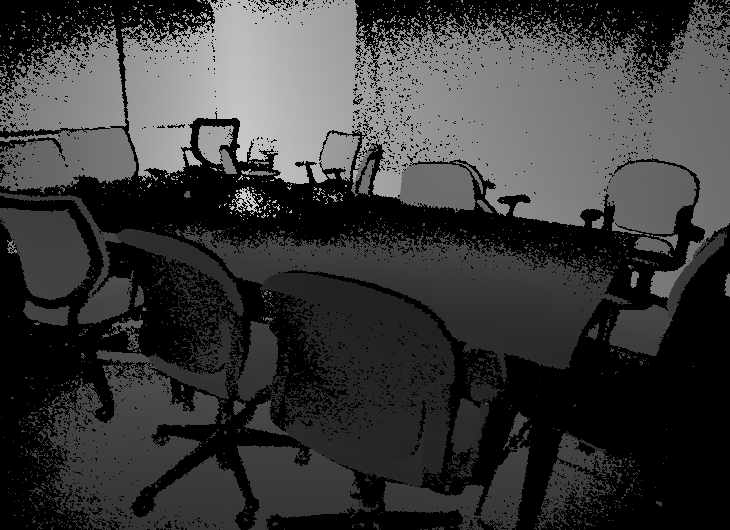} &
Depth\tabularnewline
(c) Classroom &
(d) Conference room &
\tabularnewline
\end{tabular}
\par\end{centering}

\caption{\label{fig:rgb_vs_depth} Pairs of RGB-D images from the SUN RGB-D
database. Black pixels represent regions where depth information is
not available, where the objects are too far to be captured by depth
cameras (but not by RGB cameras). This loss of depth information can
lead to category confusion in depth images, e.g., (a) \textit{furniture
store} vs (b) \textit{bedroom}, (c) \textit{classroom} vs (d) \textit{conference
room}.}
\end{figure}

A second limitation of current RGB-D scene recognition on images is
the limited range of depth cameras, in addition to less accurate information with distance. For instance, the effective range of the depth sensor
of the widely used Microsoft Kinect is 0.5m to 5.5m, with accuracy
decreasing with distance. This leads to much more limited information
and to ambiguity in the classification than in the case of RGB images (see Fig.~\ref{fig:rgb_vs_depth},
where \textit{furniture store} and \textit{classroom} can be easily
confused with \textit{bedroom} and \textit{conference room}, respectively,
due to the limited information about distant objects). In addition,
images are also limited to capture only a fraction of large scenes.
Videos can alleviate these problems by traversing the scene, and increasing
the overall coverage of visual information. Motivated by these limitations,
we introduce a new RGB-D video database for scene recognition (ISIA
RGB-D).

We take advantage of the richer depth (and RGB) information in videos
and address RGB-D video scene recognition extending the scene
recognition architecture for images with a recurrent neural network (RNN) in
order to obtain a richer spatio-temporal embeddings. Since the training
RGB-D data is limited, we propose a three-steps training procedure:
1) weakly-supervised pretraining of CNNs with depth data, 2) pretraining
of temporal embedding with frames, and 3) joint spatio-temporal fine
tuning. 

A preliminary version of this work was presented in \cite{Song_2017_AAAI},
which mainly focuses on addressing the problem of limited depth data
for RGB-D scene recognition. In this paper we extend that work to
address the problem of fully capturing depth information in wide scenes,
since depth cameras only capture depth information in a short range.
We introduce the ISIA RGB-D video database to study scene recognition
under these settings. We propose a CNN-RNN framework to model and
recognize RGB-D scenes. Inspired by the effectiveness of the two-step
training strategy in the still image case (pretraining with patches
followed by fine tuning with full images), we further propose a three-step
training procedure for the CNN-RNN architecture. Our evaluations show
significant gains obtained when integrating depth video in comparison
to still images. 

\section{Related Work}

\subsection{RGB-D scene recognition}

Earlier works use handcrafted features, engineered by experts to capture
some specific properties considered representative. Gupta \textit{et
al.} \cite{Gupta2015} propose a method to detect contours on depth
images for segmentation, then further quantize the segmentation outputs
as local features for scene classification. Banica \textit{et al.}
\cite{Banica_2015_CVPR} quantize local features with second order
pooling, and use the quantized feature for segmentation and scene
classification. More recently, multi-layered neural networks can learn features
directly from large amounts of data. Socher \textit{et al.} \cite{Socher2012}
use a single layer CNN trained unsupervisedly on patches, and combined
with a recurrent neural network (RNN). Gupta \textit{et al.}
\cite{GuptaECCV2014} use R-CNN on depth images to detect objects
in indoor scenes. Since the training data is limited, they augment
the training set by rendering additional synthetic scenes.

Current state-of-the-art relies on transferring and fine tuning Places-CNN
to RGB and depth data \cite{Gupta_2016_CVPR,Wang_2016_CVPR,Zhu_2016_CVPR,Song2015a}.
Wang \textit{et al.} \cite{Wang_2016_CVPR} extract deep features
on both local regions and whole images on both RGB, depth and surface
normals, and then use component-aware fusion to combine these multiple
components. Some approaches \cite{Zhu_2016_CVPR,Gupta_2016_CVPR}
propose incorporating CNN architectures to jointly fine tune RGB and
depth image pairs. Zhu \textit{et al.} \cite{Zhu_2016_CVPR} jointly
fine tune the RGB and depth CNN models by including a multi-modal
fusion layer, simultaneously considering inter and intra-modality
correlations, meanwhile regularizing the learned features to be compact
and discriminative. Alternatively, Gupta \textit{et al.} \cite{Gupta_2016_CVPR}
propose a cross-modal distillation approach where learning of depth
filters is guided by the high-level RGB features obtained from the
paired RGB image. Note that this method makes use of additional unlabeled
frames during distillation.

In this paper we avoid relying on large yet still RGB-specific models
to obtain depth features, and train depth CNNs directly from depth
data, learning truly depth-specific and discriminative features, compared
with those transferred and adapted from RGB models.

\subsection{Weakly-supervised CNNs}

Accurate annotations of the objects (i.e. category and bounding boxes)
in a scene are expensive and often not available. However, image-level
annotations (e.g., category labels) are cheaper to collect. These
weak annotations have been used recently in weakly supervised object
detection frameworks \cite{Durand_2016_CVPR,Bilen_2016_CVPR,Oquab_2015_CVPR}.
Oquab \textit{et al.} \cite{Oquab_2015_CVPR} propose an object detection
framework to fine tune pretrained CNNs with multiple regions, where
a global max-pooling layer selects the regions to be used in fine
tuning. Durand \textit{et al.} \cite{Durand_2016_CVPR} extend this
idea by selecting both useful (positive) and \textquotedbl{}useless\textquotedbl{}
(negative) regions with a maximum and minimum mixed pooling layer.
Bilden and Vedaldi \cite{Bilen_2016_CVPR} use region proposals to
select regions. Weakly supervised learning has been also used in RGB
scene recognition \cite{Rasiwasia2009,Rasiwasia2012,Song2015,Song2016,wang2017weakly,song2017multi}.
Similarly to the previous case, image-level labels are used to supervise
the learning of mid-level features localized in smaller regions. For
example, the classification model of \cite{Rasiwasia2012} is trained
with patches that inherit the scene label of the images. That training
process is considered as weak supervision since patches with similar
visual appearances may be assigned different scene labels.

These works often rely on CNNs already pretrained on large RGB datasets,
and weak supervision is used in a subsequent fine tuning or adaptation
stage to improve the final features for a particular task. In contrast,
our motivation is to train depth CNNs when data is very scarce, with
a weakly supervised CNN for model initialization. In particular, we pretrain
the convolutional layers prior to fine tuning with full images.

\subsection{Scene recognition on sequential data}

Previous works on scene recognition with videos focus on RGB data
\cite{Feichtenhofer_2017_CVPR,Derpanis_2012_CVPR,Shroff_2010_CVPR}.
Moving vistas \cite{Shroff_2010_CVPR} focuses on scenes with highly
dynamic patterns, such as fire, crowded highways or waterfalls, using
chaos theory to capture dynamic attributes. Derpanis \textit{et al.}
\cite{Derpanis_2012_CVPR} study how appearance and temporal dynamics
contribute to scene recognition. Feichtenhofer \textit{et al.} \cite{Feichtenhofer_2017_CVPR}
propose a new dataset with more categories and an architecture using
residual units and convolutions across time. However, none of these
video databases have depth data, and also these databases contain
outdoor natural scenes where is difficult to capture depth information.

There are some datasets involving RGB-D videos of scenes. The SUN3D
database \cite{xiao2013sun3d} contains videos of indoor scenes primarily
to study structure from motion and semantic segmentation. The NYUD2
dataset \cite{Silberman2012} contains videos with some images annotated
scene labels and with semantic segmentations. However, it contains
27 categories but only 10 are well represented, of which one or two
categories could be considered wide scenes (e.g., furniture store).
In this paper we focus more on wide scenes, which can benefit more
from traversing the scene, and propose a new dataset better suited
to study RGB-D scene recognition in videos.

\subsection{Embedding sequential data}

Recent works on visual recognition with videos incorporate spatio-temporal
deep models. Action recognition is an example that requires modeling
appearance and temporal dynamics. Many works in this area use spatio-temporal
CNN models \cite{Karpathy_2014_CVPR} or two stream combining appearance
and motion \cite{Simonyan_2014_NIPS,Feichtenhofer_2016_CVPR}. Feichtenhofer
\textit{et al.} \cite{Feichtenhofer_2017_CVPR} propose to extend
the two stream CNN with a ResNet architecture \cite{He_2016_CVPR}
and apply it to scene recognition.

\section{Depth Features from RGB Features}

Deep CNNs trained with large datasets can extract excellent representations
that achieve state-of-the-art performance in a wide variety of recognition
tasks \cite{Jeff2014}. In particular, those trained with the Places
205 database (hereinafter Places-CNN) are essential to achieve state-of-the-art
scene recognition accuracy \cite{Zhou2014}, even simply using a linear
classifier or fine tuning the parameters of the CNN. This knowledge
transfer mechanism has been used extensively (e.g., domain adaptation)
but mostly within the RGB modality (intra-modal transfer). Therefore,
it is not clear its effectivity with depth (cross-modal transfer).

Similarly to RGB features, depth features can be handcrafted or learned
from data. Since there is no large dataset of depth images, the common
approach is to transfer RGB features from deep RGB CNNs, due to certain
similarities between both modalities (see Fig.~\ref{fig:rgb_vs_depth}
and \ref{fig:motivation}).

In this section we compare intra-modal and cross-modal transfer of
a Places-CNN to RGB and depth, respectively, analyze its limitations
and explore other combinations of transfer and learning to learn better
depth features.

\subsection{Places-CNN for RGB and depth data}

We focus first on the first convolutional layer (\textit{conv1}),
since it is the closest to the input data and therefore essential
to capture modality-specific patterns.

\begin{figure}[tbh]
\begin{centering}
\includegraphics[width=0.9\columnwidth]{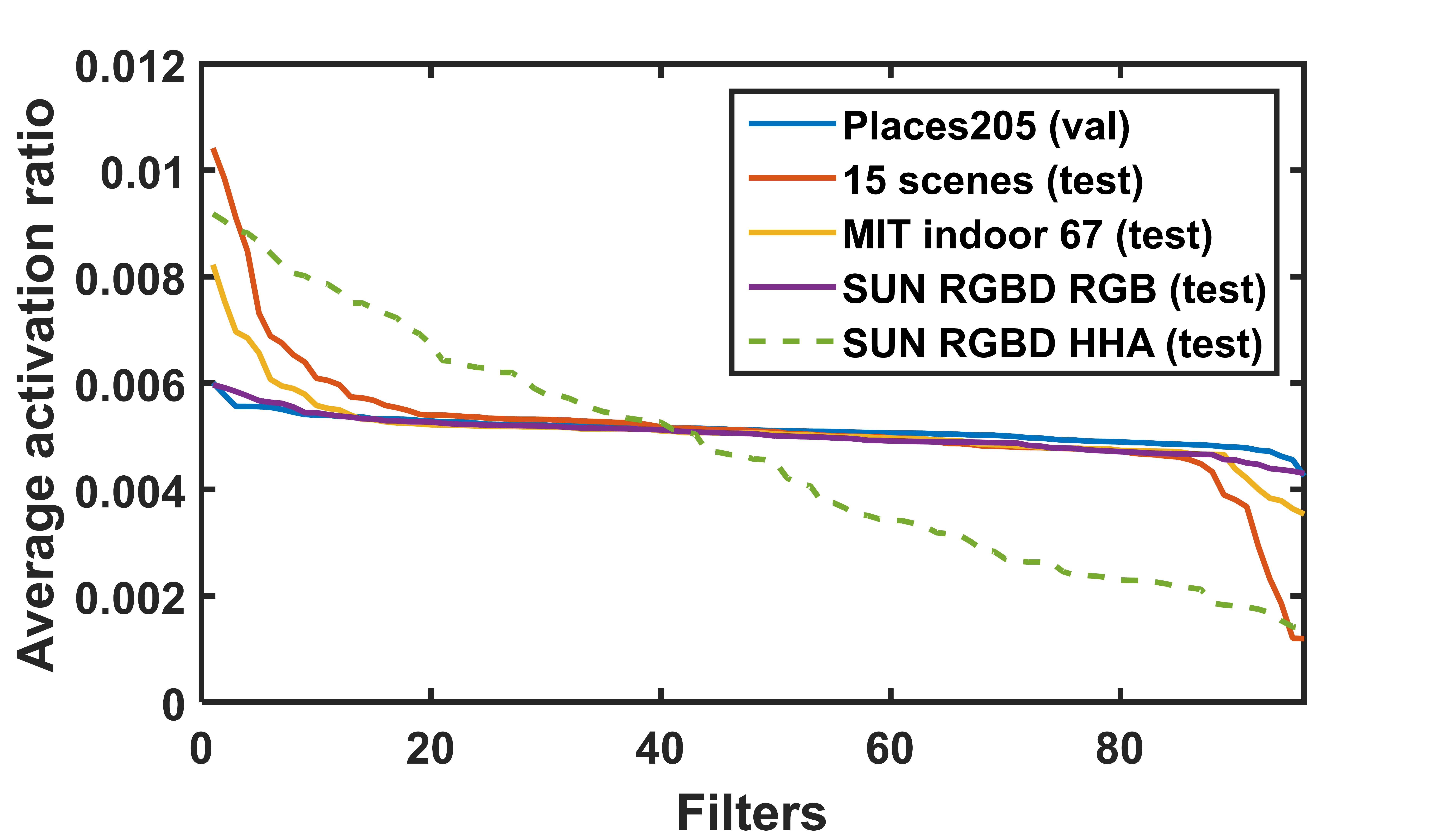} 
\par\end{centering}

\begin{centering}
\includegraphics[width=0.95\columnwidth]{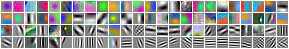} 
\par\end{centering}

\begin{centering}
\includegraphics[width=0.38\columnwidth]{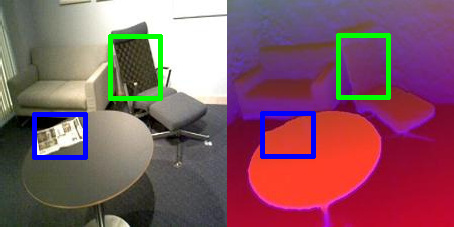}
\hskip 0.1\columnwidth \includegraphics[width=0.38\columnwidth]{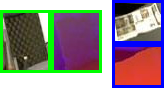}
\par\end{centering}

\caption{\label{fig:motivation}Patterns in RGB and depth modalities. Top:
average nonzero activations of the filters in the conv1 layer of Places-CNN
on different scene datasets. Middle: Conv1 filters ordered by mean
activation on SUN RGB-D HHA. Bottom: examples of scenes captured in
RGB and depth (HHA encoding) with textured regions highlighted.}
\end{figure}

Fig.~\ref{fig:motivation} (top) shows the average activation ratio
(in descending order) of the 96 filters in the layer \textit{conv1}
of a Places-CNN with AlexNet architecture \cite{Krizhevsky2012}.
Activation rate here indicates how often the response of a particular
filter is non-zero. When the input data is the validation set of Places
205 (i.e., same input distribution as in the source training set),
the curve is almost flat, showing that the network is well designed
and trained, with all the filters contributing almost equally to build
discriminative representations. When the input is from other RGB scene
datasets, such as 15 scenes \cite{Lazebnik2006}, MIT Indoor \cite{Quattoni2009}
and the RGB images from SUN RGB-D \cite{Song2015a}, the curves are
very similar, i.e., a flat activation rate for most filters and just
a few filters with higher or lower activation rate, due mostly to
the particular biases of the datasets. This shows the majority of
the filters in \textit{conv1} are good representations of the low-level
patterns in RGB scenes (see Fig.~\ref{fig:motivation} middle). This
is reasonable, since these patterns are observed in similar proportions
in both the source and target datasets.

Now let us consider the same SUN RGB-D dataset as input data, but
depth images instead of RGB (in HHA encoding, see Fig.~\ref{fig:motivation}
bottom). While still representing the same scenes, the activation
rate curve shows a completely different behavior, with only a subset
of the filters being relevant and a large number being rarely activated.
This illustrates how RGB and depth modalities are significantly different
at the low-level. In HHA encoded depth images we can still observe
edges and smooth gradients, but other patterns such as texture are
simply not present in that modality (observe in Fig.~\ref{fig:motivation}
bottom how the textures in the newspaper and the chair back completely
disappear in HHA images). This can be observed more clearly by rearranging
the filters according to decreasing activation rate with HHA images
(see Fig.~\ref{fig:motivation} middle), and see how the most frequently
activated filters are typically those dealing with smooth color variations
and edges (yet not optimal), while the least activated deal with RGB-specific
features such as Gabor-like and high frequency patterns. 

\subsection{Fine tuning with depth data}

The previous result suggests that adapting bottom layers is more important
when transferring to depth. In previous works \cite{Song2015a,Wang_2016_CVPR,Zhu_2016_CVPR,wang2016correlated,eitel2015multimodal}
the depth network is fine tuned only in the top layers as a whole,
but with such limited data it will still have difficulty to reach
and properly adapt the bottom layers. In contrast, we want to emphasize
explicit adaptation in bottom layers, since they are more critical
to capture modality-specific patterns.

\begin{figure}[tbh]
\begin{centering}
\includegraphics[width=1\columnwidth]{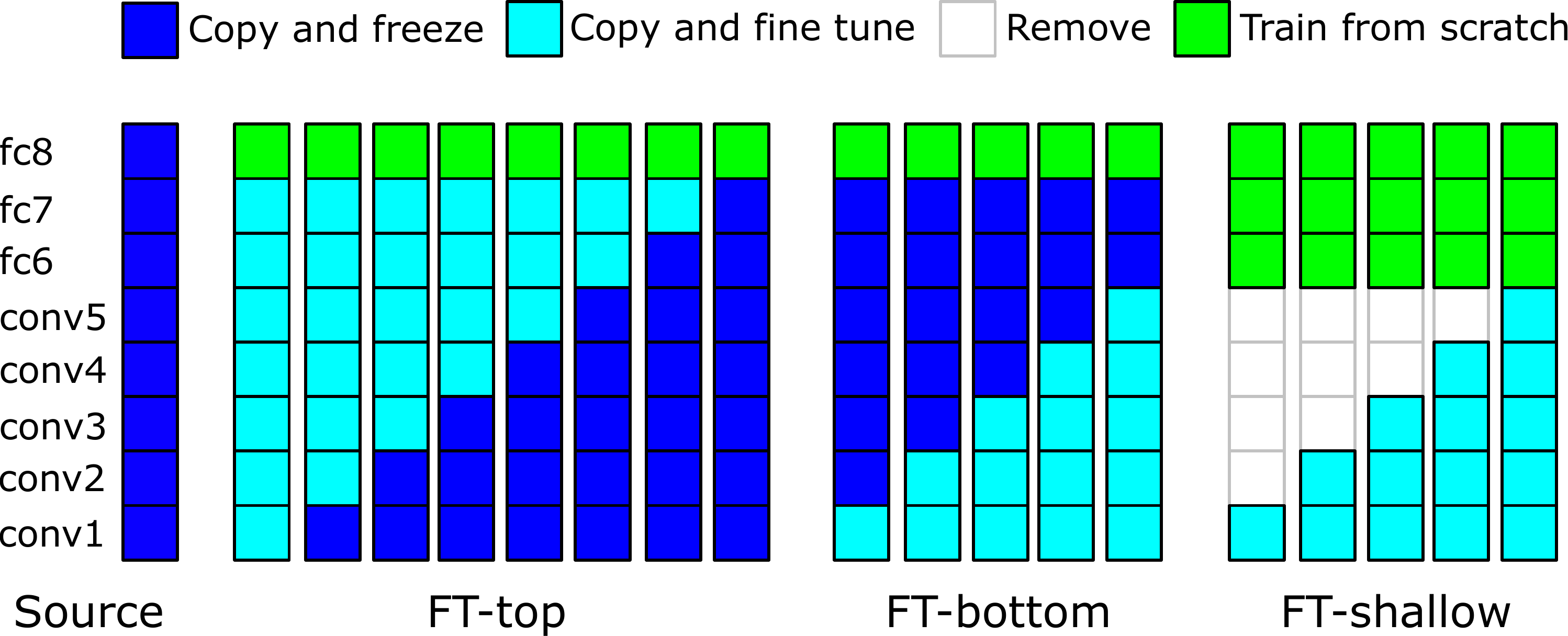} 
\par\end{centering}

\caption{\label{fig:ft-strategy} Different settings for learning depth features,
combining transfer from a source Places-CNN (AlexNet architecture),
fine tuning and training from scratch: (a) top layers, (b) bottom
layers, and (c) bottom layers with some intermediate convolutional
layers removed. Each column represents a particular setting.}
\end{figure}

Here we explore alternatives to learn depth representations, considering
several factors: parameter initialization and tuning (trained from
scratch, fine tuned or frozen), the position of the trainable/tunable
layers (top or bottom), and the overall depth of the network. With
this in mind, we organize these settings in three groups (see Fig.~\ref{fig:ft-strategy},
where each column represents a particular setting): a) \textit{FT-top}:
the conventional method where only a few layers at the top are fine
tuned, b) \textit{FT-bottom}: where a few layers at the bottom are
fine tuned, and c) \textit{FT-shallow}: a few convolutional layers
are kept and fine tuned while the others are removed. Note that \textit{fc8}
is always trained, since it must be resized according to the target
number of categories. In \textit{FT-shallow} we also train the other
two fully connected layers.

\begin{figure}[tbh]
\begin{centering}
\includegraphics[bb=140bp 0bp 2030bp 1077bp,clip,width=1\columnwidth]{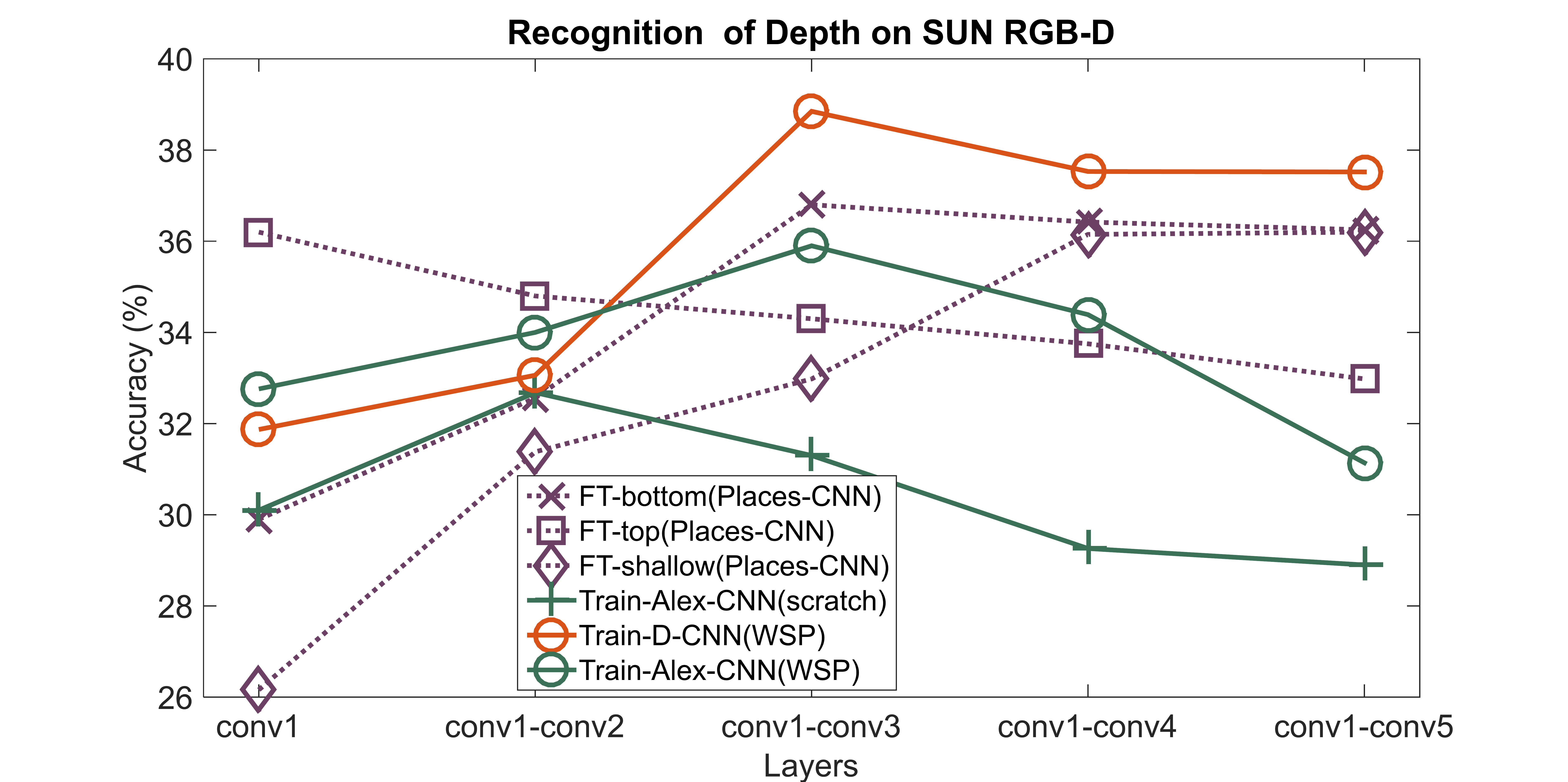} 
\par\end{centering}

\caption{\label{fig:Comparisons-between-different} Comparison of different
fine tuning and training strategies evaluated on SUN RGB-D (depth
images). The figure includes curves for methods based on fine tuning
Places-CNN (see Fig.~\ref{fig:ft-strategy}) and for methods based
on training from scratch (see Fig.~\ref{fig:tfs-strategy}). The
horizontal axis shows the number of convolutional layers being trained
(\textit{scratch}) or fine tuned. }
\end{figure}

The classification accuracy on the depth data of the SUN RGB-D dataset
is shown Fig.~\ref{fig:Comparisons-between-different}. We first
analyze the three FT-curves (i.e. \textit{FT-bottom}, \textit{FT-keep},
\textit{FT-WSP} and \textit{FT-top}, also see Fig.~\ref{fig:ft-strategy}),
obtained by transferring and fine tuning Places-CNN. Fine tuning top
layers (\textit{FT-top}) does not help significantly until including
bottom convolutional layers, which contrasts with RGB where fine tuning
one or two top layers is almost enough to reach the maximum gain \cite{Pulkit_2014}.
Further extending fine tuning to bottom layers in RGB helps very marginally.
This agrees with the previous observation that bottom layers and \textit{conv1}
in particular need to be adapted to the corresponding modality. In
fact, fine tuning only the three bottom layers (\textit{FT-bottom})
achieves 36.5\% accuracy which is higher than fine tuning the whole
network, probably due to overfitting. We also evaluated shallower
networks with fewer convolutional layers and therefore fewer parameters
(\textit{FT-shallower}), where we observe again that fine tuning the
first layers contributes most to high accuracy. Again, these results
suggest that adapting bottom layers is much more important when transferring
RGB to depth, and therefore that fine tuning for intra-modal and cross-modal
transfer should be handled differently.

\subsection{More insight from layer conv1}

\begin{figure}[tbh]
\begin{centering}
\begin{tabular}{cc}
\includegraphics[width=0.4\columnwidth]{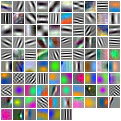}  &
\includegraphics[width=0.4\columnwidth]{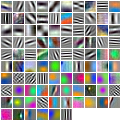}\tabularnewline
(a)  &
(b)\tabularnewline
\includegraphics[width=0.4\columnwidth]{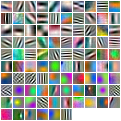}  &
\includegraphics[width=0.4\columnwidth]{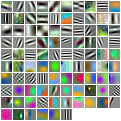}\tabularnewline
(c)  &
(d)\tabularnewline
\includegraphics[width=0.4\columnwidth]{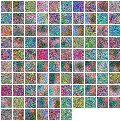}  &
\includegraphics[width=0.4\columnwidth]{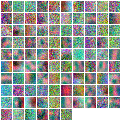}\tabularnewline
(e)  &
(f)\tabularnewline
\includegraphics[width=0.4\columnwidth]{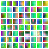}  &
\includegraphics[width=0.4\columnwidth]{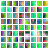}\tabularnewline
(g)  &
(h)\tabularnewline
\end{tabular}
\par\end{centering}

\caption{\label{fig:Visual_conv1}Visualizing the first convolutional layer
(conv1): (a) Places-CNN; (b) full fine tuned Places-CNN; (c) FT-bottom
(Places-CNN); (d) FT-shallower (Places-CNN), conv1; (e) Train-Alex-CNN
(scratch); (f) Train-Alex-CNN (WSP), training with patches ($99\times99$
pixels); (g) WSP-CNN, kernel size $5\times5$ pixels, training with
patches ($35\times35$ pixels); (h) Train-D-CNN (WSP). All methods
are trained/fine tuned using only the depth data from SUN RGB-D. }
\end{figure}

We can compare the filters obtained in \textit{conv1} with these different
settings for additional insight (see Fig.~\ref{fig:Visual_conv1}).
Although there is some gain (when fine tuning) in accuracy, only a
few particular filters have noticeable changes during the fine tuning
process (see Fig.~\ref{fig:Visual_conv1} from (a) to (d)). This
suggests that the CNN is still mainly reusing the original RGB filters,
and thus trying to find RGB-like patterns in depth data. As Fig.~\ref{fig:motivation}
middle shows, a large number of filters from Places-CNN are significantly
underused on depth data (while they are properly used on RGB data).
These observations suggest that reusing Places-CNN filters for \textit{conv1}
and other bottom layers may not be a good idea. Moreover, since filters
also represent tunable parameters, this results in a model with too
many parameters that is difficult to train with limited data.

\section{Learning effective depth features}

In the previous section, it can be observed that transferring and
fine tuning Places-CNN with depth data is somewhat effective but limited
to exploiting some specific RGB-like patterns in depth images. It
also can be observed that bottom layers seem to be the most important
when learning modality-specific features. Here we aim at learning depth-specific
features of early convolutional layers, directly from depth data,
that are at least competitive in performance with those transferred
from Places-CNN. The main problem is the limited depth data and the
complexity of Places-CNN (i.e., large number of parameters).

\subsection{Weak supervision on patches}

We propose to work on patches instead of full images and adapt the
complexity of the network to accommodate smaller feature maps and
the amount of training data. In this sense we can increase the training
data while reducing the number of parameters, which will help to learn
discriminative filters in the first convolutional layers. Since patches
typically cover objects or parts of objects, in principle the original
scene labels are not suitable for supervision. For instance, the scenes
categories \textit{living room, dining room, }and \textit{classroom}
often contain visually similar patches since they may represent mid-level
concepts such as \textit{walls, ceilings, tables }and \textit{chairs},
but suitable mid-level labels or object annotations are not available.
However, a particular patch can be weakly labeled with the corresponding
scene category of the image. This weak supervision has been proved
helpful to learn discriminative features for RGB scene recognition \cite{Rasiwasia2009,song2015joint,Song2016,song2017multi}.
Hence, we refer to this network as \textit{weakly supervised patch-CNN}
(WSP-CNN). Once the network is trained, the parameters of the WSP-CNN
are used to initialize the convolutional layers of the full network,
which is then fine tuned with full images.

\begin{figure}[tbh]
\begin{centering}
\includegraphics[width=1\columnwidth]{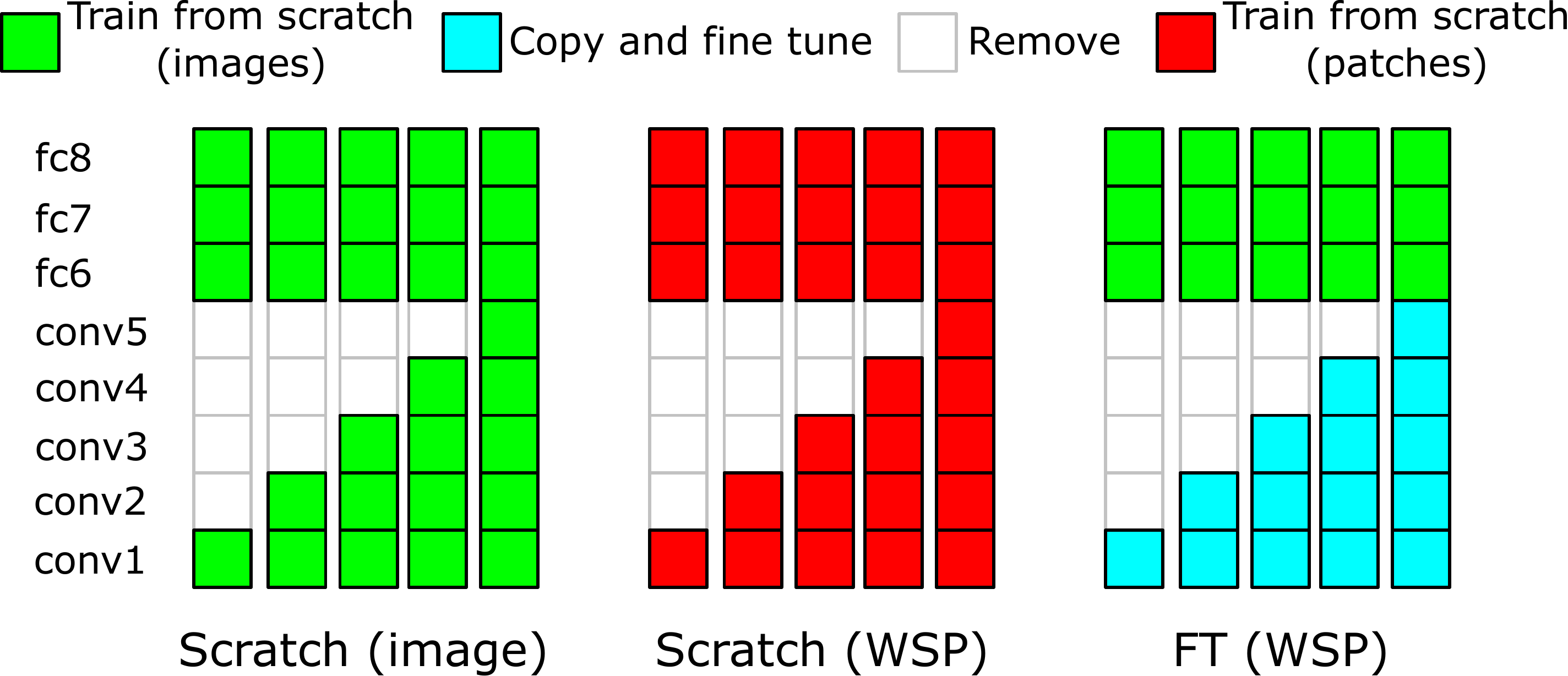}
\par\end{centering}

\caption{\label{fig:tfs-strategy}Training strategies for Alex-CNN variants
with depth images, (a) from scratch, (b) weakly-supervised with patches,
and (c) fine-tuned after weakly supervised training with patches.}
\end{figure}

We first implement this strategy on the AlexNet architecture (hereinafter
Alex-CNN). We sample a grid of $4\times4$ patches of $99\times99$
pixels for weakly-supervised pretraining. When switching from WSP-CNN
to Alex-CNN, only the weights of the convolutional layers are transferred.
Fig.~\ref{fig:Comparisons-between-different} shows that using this
pretraining stage significantly outperforms training directly with
full images (compare \textit{Train-Alex-CNN (WSP)} vs \textit{Train-Alex-CNN
(scratch)}). Furthermore, in the \textit{conv1} filters shown in Fig.~\ref{fig:Visual_conv1}f
(WSP) the depth specific-patterns are much more evident than in Fig.~\ref{fig:Visual_conv1}e
(full image). Nevertheless, they still show a significant amount of
noise (probably remnant of the original random initialization which
still cannot vanish with such limited training data). This suggests
that AlexNet is still too complex, and perhaps the size of Alex-CNN
kernels may be too large for depth data.

\begin{figure*}[!t]

\begin{centering}
\includegraphics[width=0.9\textwidth]{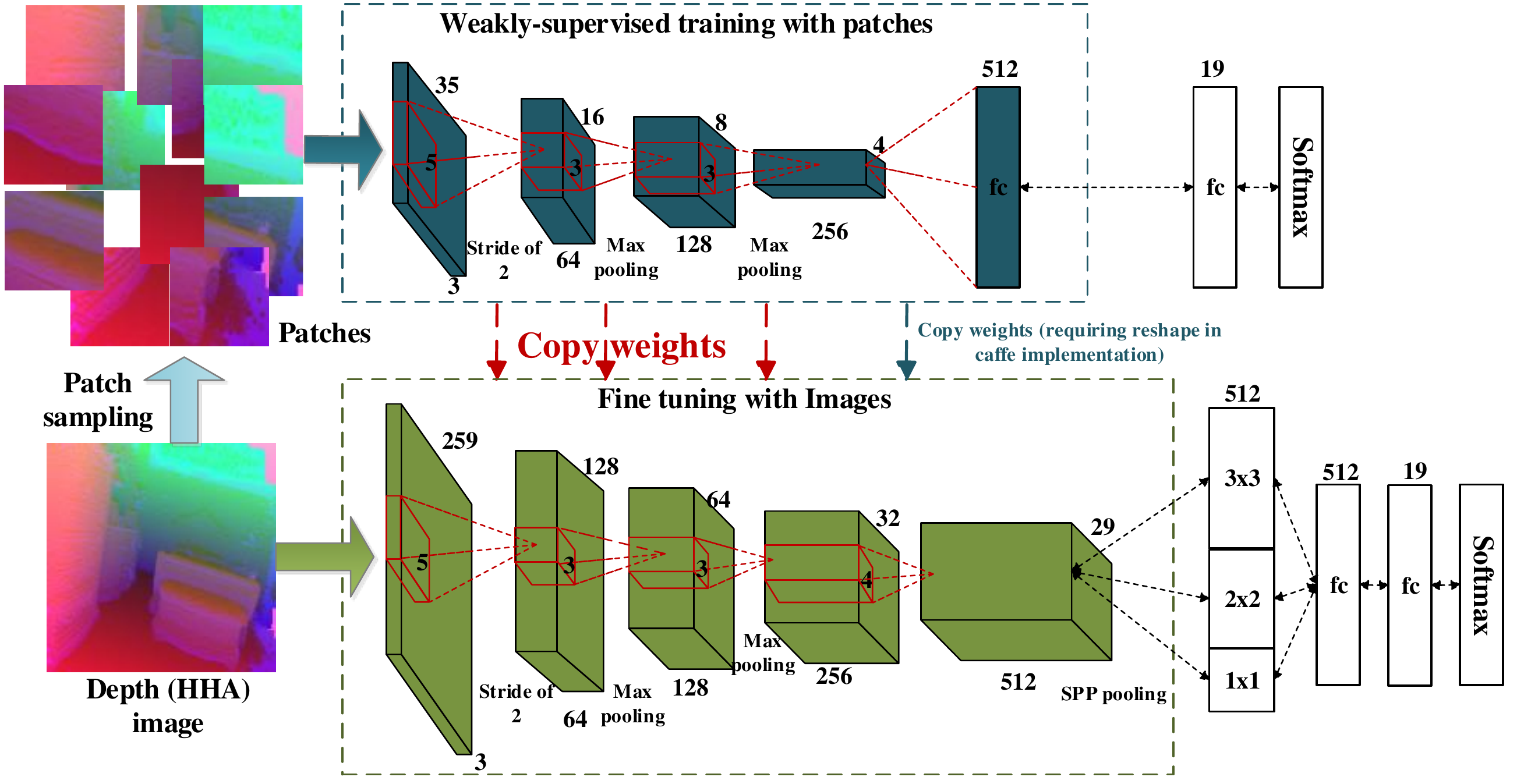} 
\par\end{centering}

\caption{\label{fig:RGB-D-scene-recognition}Two-step learning of depth CNNs
combining weakly supervised pretraining and fine tuning.}
\end{figure*}

\subsection{Interpretation as category co-occurrence modeling}

Our approach can be seen as a first set of layers that learns an intermediate
local semantic representation (e.g., objects, regions), while the
other set of layers corresponds to a second model that infers scenes
from a set of intermediate representation. This actually resembles
earlier two-step local-to-global approaches to scene recognition using
intermediate representations (e.g., bag-of-words, topic models).

In particular, a weakly supervised model followed by a global model
resembles previous works on scene category co-occurrence modeling
\cite{Rasiwasia2009,Song2016}. Supervised by global labels, the model
can predict scene categories directly (e.g., pooling the outputs of
the softmax) \cite{Rasiwasia2009}. However, the weak supervision
with scene categories makes the prediction very ambiguous, resulting
in visually related categories predicted with similar probabilities
due to lack of global context. Luckily, these co-occurrence patterns
are consistent across categories, so the second model exploits them
to resolve the ambiguity \cite{Song2016}, often combined with spatial
and multi-feature contexts \cite{song2015joint,song2017multi}.

In contrast to previous works about category co-occurrences, we do
not use probabilities as intermediate representations, but the activations
before the softmax. This makes training easier. In general, all layers
in deep networks can be trained jointly, as long as the training data
is enough. However, when training data is limited, this two-step procedure
with weak supervision seems to be very helpful.

\subsection{Depth-CNN}

Since the complexity and diversity of patterns found in depth images
are significantly lower than those found in RGB images (e.g., no textures),
we reduced the number of convolutional layers to three and also the
size of the kernels in each layer (see Fig~\ref{fig:RGB-D-scene-recognition}
top for the details). The sizes of the kernels are $5\times5$ (stride
2), $3\times3$ and $3\times3$, and the size of max pooling is $2\times2$,
stride 2. We sample a grid of $7\times7$ patches of $35\times35$
pixels for weakly-supervised pretraining.

Fig~\ref{fig:RGB-D-scene-recognition} bottom shows the full architecture
of the proposed depth-CNN (D-CNN). After weakly supervised pretraining,
we transfer the weights of the convolutional layers. The output of
\textit{conv4} in D-CNN is $29\times29\times512$, almost 50 times
larger than the output of pool5 (size of $6\times6\times256$) in
Alex-CNN, which leads to 50 times more parameters in this part. In
order to reduce the number of parameters in the next fully connected
layer, we include a spatial pyramid pooling (SPP) \cite{He_2014}
composed of three pooling layers of size of $29\times29$,$15\times15$,
$10\times10$. SPP also captures spatial information and allows us
to train the model end-to-end. This model outperforms Alex-CNN, both
fine tuned and weakly-supervised trained (see \textit{D-CNN (WSP)}
in Fig.~\ref{fig:Comparisons-between-different}). Comparing the
visualizations in Fig.~\ref{fig:Visual_conv1}, the proposed WSP-CNN
and D-CNN learn more representative kernels, which also helps to improve
the performance. This also suggests that smaller kernel sizes are
more suitable for depth data since high frequency patterns requiring
larger kernels are not characteristic of this modality.

\section{Multimodal RGB-D architecture}

Most previous works use two independent networks for RGB and depth,
that are fine tuned independently, then another stage exploits correlation
between RGB and depth features and finally another stage learns the
classifier \cite{Zhu_2016_CVPR,Wang_2016_CVPR,Wang2015}. This stages
are typically independent. In contrast, we integrate both RGB-CNN,
depth-CNN and the fusion procedure into an integrated RGB-D-CNN, which
can be trained end-to-end, jointly learning the fusion parameters
and fine tuning both RGB layers and depth layers of each branch. As
fusion mechanism, we use two fully connected layers followed by the
loss, on top of the concatenation of RGB and depth features.

Recent works exploit metric learning \cite{Wang2015}, Fisher vector
pooling \cite{Wang_2016_CVPR} and correlation analysis \cite{Zhu_2016_CVPR}
to reduce the redundancy in the joint RGB-D representation. It is
important to note that this step can improve the performance significantly
when RGB and depth features are more correlated. This is likely to
be the case in recent works when both RGB and depth feature extractors
are fine tuned versions of the same CNN model, as we saw in previous
sections with Places-CNN. In our case depth models are learned directly
from depth data and independently from RGB, so they are already much
less correlated, and therefore those methods are not so effective
in our case and a simple linear fusion layer works just fine. 
\begin{figure*}[t]
\centering{}\setlength{\tabcolsep}{1pt} \global\long\def\arraystretch{1}
\begin{tabular}{c}
\includegraphics[width=1\textwidth]{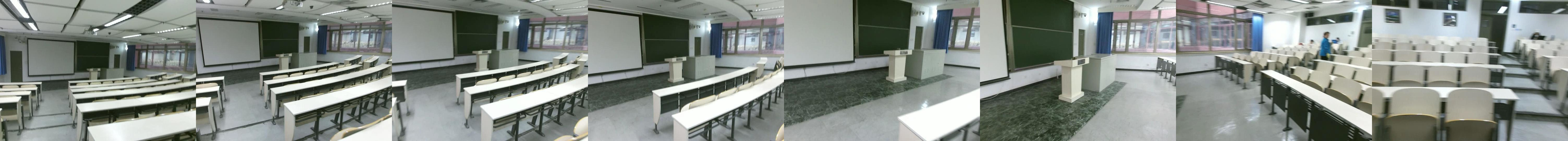} \tabularnewline
\includegraphics[width=1\textwidth]{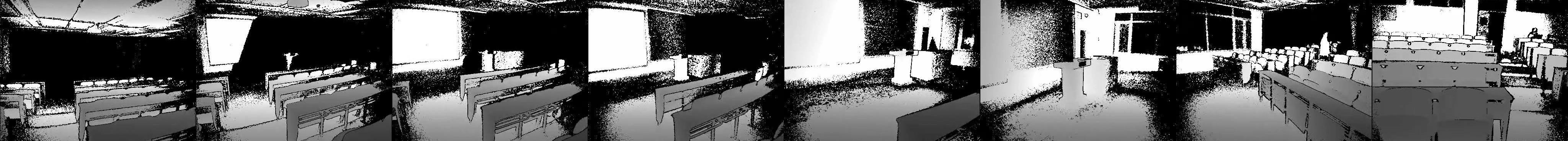}
\end{tabular}\caption{\label{fig:The-process} Capturing process of a \textit{classroom}
scene. Note that this wide and extend scene requires more footage
than other cases.}
\end{figure*}

\begin{figure*}[tbh]
\begin{centering}
\includegraphics[bb=20bp 0bp 2120bp 697bp,clip,width=1\textwidth]{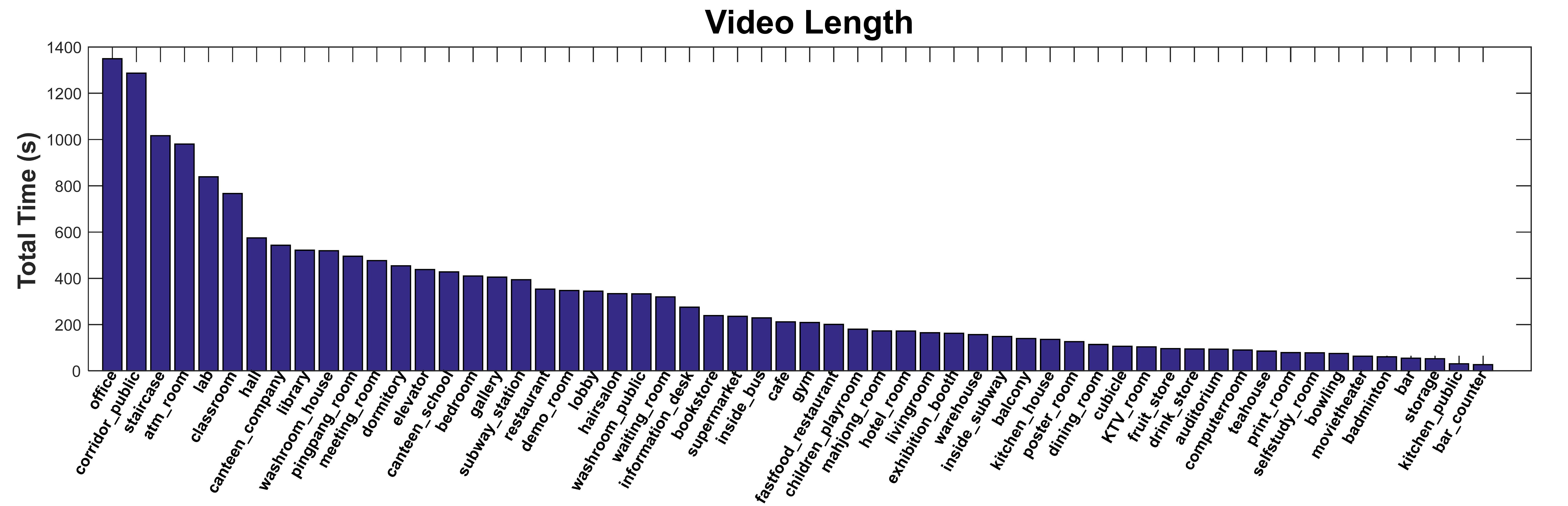} 
\par\end{centering}

\caption{\label{fig:Overview-database} Distribution of scene categories in
ISIA RGB-D.}
\end{figure*}

\section{ISIA RGB-D Video Database}

In order to investigate scene recognition in RGB-D videos, we introduce
the ISIA RGB-D video database\footnote{This database is released in the following link: http://isia.ict.ac.cn/dataset/ISIA-RGBD.html}.
It contains indoor videos captured from three different cities (separated
up to 1000 km), guaranteeing diversity in locations and scenes. The
database reuses 58 of the categories in the taxonomy of the MIT indoor
scene database \cite{Quattoni2009}, and has a total of 278 videos,
with more than five hours of footage in total. The duration of the
footage per category is shown in Fig.~\ref{fig:Overview-database}.
The duration of videos varies, depending on the complexity and extension
of the scene itself (a \textit{classroom} or \textit{furniture store}
requires more footage than \textit{office} or \textit{bedroom}) and
how common and easy to access are certain categories (e.g., \textit{office}
and \textit{classroom} have more videos than \textit{auditorium} or
\textit{bowling alley}). Videos are captured using a Microsoft Kinect
version 2 sensor, with a frame rate of 15 frames/s, obtaining more
than 275000 frames.

The database aims at addressing the limitations of the narrow field
of view in conventional RGB-D sensors and the limited range of the
depth one, by increasing the coverage by recording videos instead
of images. In particular, it targets wide scenes, which we capture
by starting on one side and moving to the other across the scene while
panning the camera to maximize the coverage. Fig.~\ref{fig:The-process}
shows an example of the category \textit{classroom}). Note that regions
like the podium, the whiteboard and the windows are missing in the
initial depth image, but are captured in other parts of the video
sequence.

\section{CNN-RNN Architecture for Video Recognition}

\begin{figure}[tbh]
\begin{centering}
\includegraphics[bb=0bp 0bp 973bp 572bp,width=1\columnwidth]{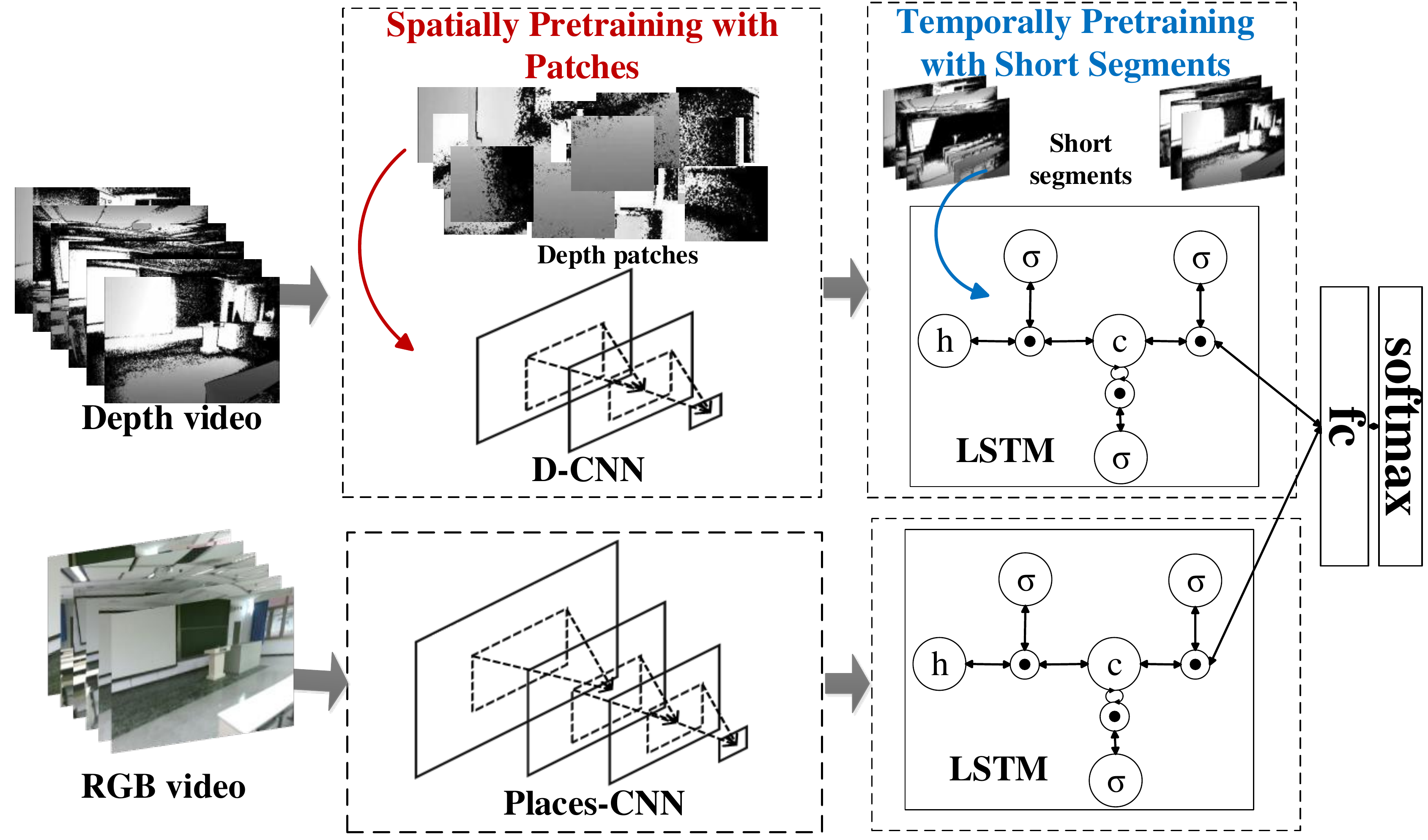} 
\par\end{centering}

\caption{\label{fig:Framework.}Framework of RGB-D video recognition.}
\end{figure}

Scene recognition with video data requires aggregating spatial features
along time into a joint spatiotemporal representation. We propose
a framework combining convolutional and recurrent neural networks
that would capture spatial and temporal information, respectively,
in a joint embedding (see Fig.~\ref{fig:Framework.}). Particularly,
the recurrent neural networks are implemented using Long-Short Term
Memory (LSTM) units.

Similar to other works, our framework has independent branches for
RGB and depth data, following the architecture for images. Temporal
embedding with LSTMs is also modality-specific, and then late fusion
is performed at sequence level using a fully connected layer. The
combined architecture is trained jointly end-to-end.

\subsection{Depth feature learning using pretraining with patches}

As in RGB-D images, we face the problem of limited data to learn good
neural representations from scratch, specially for depth data. We
use the two-step training strategy described previously to learn depth
features from limited data, in this case applied to both spatial and
temporal dimensions. For the CNN we use patches and for the RNN we
use short segments of frames, in both cases supervised by the scene
label assigned to the whole video.

We sample the video and resize the keyframes to $256\times256$ pixels.
Following a similar strategy, we first pretrain the depth CNN model
using patches. In the case of videos, patches are sampled
from keyframes, and therefore are not limited to one image but to
a segment or the whole sequence. 

The pretrained model is then fine tuned with individual keyframes
with stronger supervision. The fine tuned model can separately predict
the scene probability for each frame.

\subsection{Integrating temporal information}

We use two strategies to integrate temporal information: average pooling
and LSTM. The former is deterministic and used as baseline for comparison.
We simply average the scene probabilities (i.e., output of the softmax
of the image model) of all the keyframes. The latter exploits recurrent
relations between keyframes by learning an LSTM embedding.

\subsection{Training the temporal embedding}

In contrast to averaging, the LSTM embedding needs to be trained.
Since we have a limited number of sequences and keyframes, we follow
a similar pretraining strategy using short segments of keyframes.
In this case we apply it to both RGB and depth branches separately.

For each video, we sample sets of short segments. All these short
segments have the same length of $T$ keyframes, from which modality-specific
CNN features $X=[x_{1},x_{2},\ldots,x_{T}]$ are extracted. The core
of the LSTM architecture is the memory cell $c$, which stores knowledge
at each iteration according to the observed inputs and the current
state. The behavior of the cell is determined by three different gates
(input gate $i$, forget gate $f$, and output gate $o$) and several
gating, update and output operations

\begin{eqnarray}
i_{t} & = & \sigma\,(W_{ix}x_{t}+\,W_{im}m_{t-1}),\\
f_{t} & = & \sigma\,(W_{fx}x_{t}+\,W_{fm}m_{t-1}),\\
o_{t} & = & \sigma\,(W_{ox}x_{t}+\,W_{om}m_{t-1}),\\
c_{t} & = & f_{t}\,\odot c_{t-1}+i_{t}\odot h(W_{cx}x_{t}+\,W_{cm}m_{t-1}),\\
m_{t} & = & o_{t}\,\odot c_{t},
\end{eqnarray}
where $\sigma(\cdot)$ is the sigmoid function, $\odot(\cdot)$ represents
the product with a gate value, and $h(\cdot)$ denotes the hyperbolic
tangent function. The variable $m_{t}$ is the hidden state, $x_{t}$
is the input (CNN feature of each frame) of each step and the different
$W$ are the weight matrices of the model.

\subsection{RGB-D Fusion}

Once the modality-specific branches are pretrained, we fine tune the
joint model with the full videos. Note that, as in the case of images,
the depth CNN is pretrained with patches while the RGB CNN is pretrained
using a Places-CNN.

Although the predictions for RGB and depth can also be averaged, we
find that it is more effective to combine them using a fully connected
layer. This also allows us to train the model end-to-end.

\section{Experiments}

\subsection{Settings}

\subsubsection{Datasets}

We first evaluate scene recognition on images, comparing the proposed
D-CNN and RGB-D CNN models in two RGB-D datasets: NYUD2 \cite{Silberman2012}
and SUN RGB-D \cite{Song2015a}. The former is a relatively small
dataset with 27 indoor categories, but only a few of them are well
represented. Following the split in \cite{Silberman2012}, all 27
categories are reorganized into 10 categories, including the 9 most
common categories and an \textit{other} category consisting of the
remaining categories. The training/test split is 795/654 images. SUN
RGB-D contains 40 categories with 10335 RGB-D images. Following the
publicly available split in \cite{Song2015a,Wang_2016_CVPR}, the
19 most common categories are selected, consisting of 4,845 images
for training and 4,659 images for test.

We also evaluate the proposed video recognition method on the ISIA
RGB-D video database. Eight scene categories contain only one video,
so we use the other 50 categories (each of them with different numbers
of videos, see Fig.~\ref{fig:Overview-database}). We randomly select
nearly 60\% of the data of each category for training, while the remaining
are used for test. Following \cite{Song2015a}, we report the mean
class accuracy for evaluations and comparisons.

\subsubsection{Classifier}

Since we found that training linear SVM classifiers with the output
of the fully connected layer increases performance slightly, all the
following results are obtained including SVMs, unless specified otherwise.
\begin{itemize}
\item (wSVM): this variant uses category-specific weights during SVM training
to compensate the imbalance in the training data. The weight $w=\{w_{1}...w_{K}\}$
of each category $k$ is computed as $w_{k}=\left(\frac{\min_{i\in K}N_{i}}{N_{k}}\right)^{p}$,
where $N_{k}$ is the number of training images of the $k_{th}$ category.
We selected $p=2$ empirically by cross-validation in a preliminary
experiment. 
\end{itemize}

\subsubsection{Evaluation metric}

Following \cite{Song2015a,Wang_2016_CVPR}, we report the average
precision over all scene classes for all datasets.

\subsection{SUN RGB-D}

\subsubsection{Depth features}

We first compare D-CNN and Alex-CNN on the depth data of SUN RGB-D.
The outputs of the different layers are used as features to train
the SVM classifiers. Table~\ref{tab:Comparisons-of-different} compares
five different models. In Alex-CNN, we use all the layers of Places-CNN
to initialize the network and then fine tune it, but only the three
bottom convolutional layers when we train it from scratch, since the
performance is higher than with the full architecture (see~Fig.\ref{fig:Comparisons-between-different}).

The features extracted from the bottom layers (\textit{pool1} to \textit{conv3})
trained from scratch obtain better performance for classification
than those transferred from Places-CNN and fine tuned, even though
for the top layers is worse. Using weakly-supervised training on patches
(WSP), the performance increase is comparable to that of the top layer
of the fine tuned Places-CNN and better than that of bottom layers,
despite having fewer layers and not relying on Places data.
D-CNN consistently achieves the best performance, despite being a
smaller model.

\begin{table}[tbh]
\caption{\label{tab:Comparisons-of-different} Ablation study for different
depth models. Accuracy (\%) on SUN RGB-D.}

\centering{}\centering{}%
\begin{tabular}{c|cc|cc|c}
\hline 
Arch.  &
\multicolumn{4}{c|}{Alex-CNN} &
D-CNN\tabularnewline
\hline 
\multirow{2}{*}{Weights } &
\multicolumn{2}{c|}{Places-CNN} &
\multicolumn{2}{c|}{Scratch} &
Scratch\tabularnewline
\cline{2-6} 
 & No FT &
FT  &
No WSP  &
WSP  &
WSP\tabularnewline
\hline 
Layer  &
 &
 &
 &
 &
\tabularnewline
pool1  &
17.2  &
20.3  &
22.3  &
23.5  &
\textbf{25.3}\tabularnewline
pool2  &
25.3  &
27.5  &
26.8  &
30.4  &
\textbf{33.9}\tabularnewline
conv3  &
27.6  &
29.3  &
29.8  &
\textbf{35.1}  &
34.6\tabularnewline
conv4  &
29.5  &
32.1  &
-  &
-  &
\textbf{38.3}\tabularnewline
pool5  &
30.5  &
35.9  &
-  &
-  &
-\tabularnewline
fc6  &
30.8  &
36.5  &
30.7  &
36.1  &
-\tabularnewline
fc7  &
30.9  &
37.2  &
32.0  &
36.8  &
\textbf{40.5}\tabularnewline
fc8  &
-  &
37.8  &
32.8  &
37.5  &
\textbf{41.2}\tabularnewline
\hline 
\end{tabular}
\end{table}

We also compare to related works using only depth features (see Table~\ref{tab:Comparisons-on-only}).
For a fair comparison, we also implemented SPP for Places-CNN. D-CNN
outperforms FT-Places-CNN+SPP by 3.5\%. Using the weighted SVM both
models further improve more than 1\%.

\begin{table}[tbh]
\caption{\label{tab:Comparisons-on-only} Accuracy on SUN RGB-D with depth
features (\%)}
\centering{}%
\begin{tabular}{ccc}
\hline 
\multirow{1}{*}{}  &
\multirow{1}{*}{Method}  &
\multicolumn{1}{c}{Acc.(\%)}\tabularnewline
\hline 
\multirow{2}{*}{Proposed}  &
D-CNN  &
41.2\tabularnewline
 &
D-CNN (wSVM)  &
\textbf{42.4}\tabularnewline
\hline 
\multirow{4}{*}{State-of-the-art}  &
R-CNN+FV\cite{Wang_2016_CVPR}  &
34.6\tabularnewline
 &
FT-PL\cite{Wang_2016_CVPR}  &
37.5\tabularnewline
 &
FT-PL+SPP  &
37.7\tabularnewline
 &
FT-PL+SPP (wSVM)  &
38.9\tabularnewline
\hline 
\multicolumn{3}{c}{FT: Fine tuned, PL: Places-CNN}\tabularnewline
\end{tabular}
\end{table}

\begin{figure}[tbh]
\begin{centering}
\includegraphics[bb=46bp 180bp 550bp 590bp,clip,width=1\columnwidth]{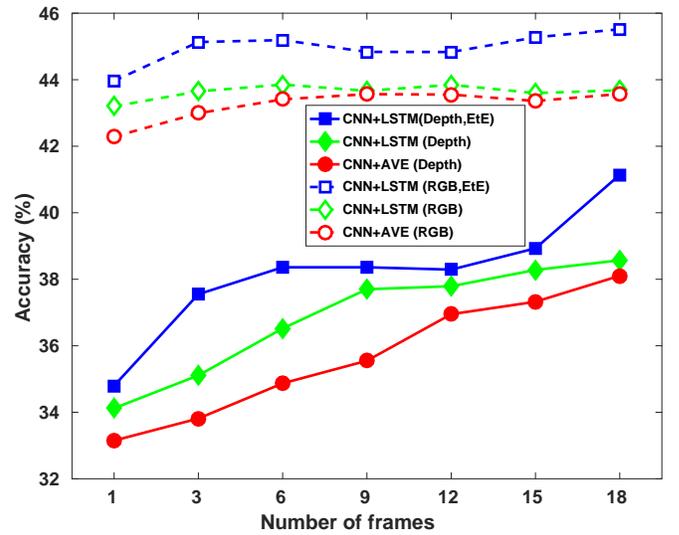} 
\par\end{centering}

\caption{\label{fig:Comparison-temporal-embedding} Impact of video length
on modality-specific recognition on ISIA RGB-D.}
\end{figure}
\begin{table}[tbh]
\caption{\label{tab:Evaluation-of-object}Evaluation of object detection with
depth data on SUN RGB-D in mAP (\%)}

\centering{}%
\begin{tabular}{c|c|c}
\hline 
Architecture &
Init. &
mAP (\%)\tabularnewline
\hline 
ZF-net \cite{Zeiler2014} &
Scratch &
29.3\tabularnewline
ZF-net \cite{Zeiler2014} &
RGB-ZF-net &
35.8\tabularnewline
\hline 
D-CNN &
Scratch &
34.6\tabularnewline
D-CNN &
WSP &
\textbf{38.1}\tabularnewline
\hline 
\end{tabular}
\end{table}

\begin{figure*}[tbh]
\begin{centering}
\includegraphics[width=0.95\textwidth]{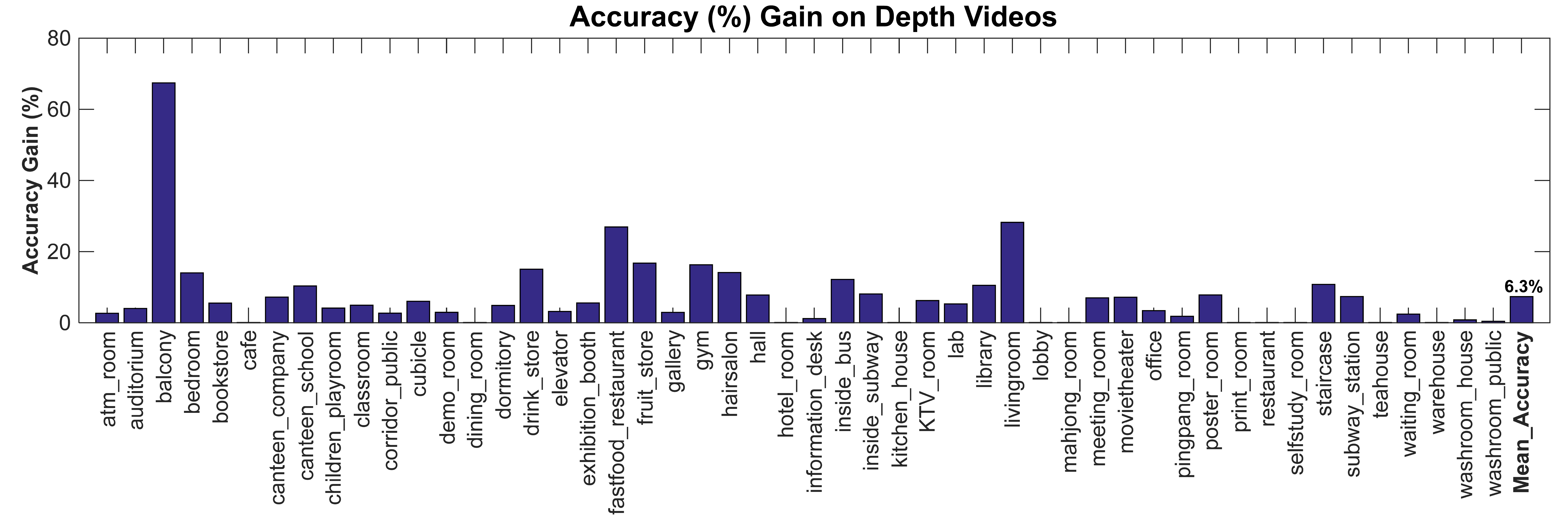}
\par\end{centering}

\caption{\label{fig:Accuracy-(=000025)-increments}Accuracy (\%) gain of using
depth video data compared to image data.}
\end{figure*}

In addition to scene recognition, we adapted the D-CNN model for object
detection. We first pretrained the CNN model with patches sampled
from depth images, which is then fine tuned with object annotations
with the framework of Faster R-CNN \cite{NIPS2015_5638}. Thus, D-CNN
is used to initialize the convolutional layers of Faster
R-CNN for object detection with depth data. Particularly for object
detection, we use object annotations (provided by SUN RGB-D) to sample
patches inside the bounding boxes of objects. The results are shown
in Figs~\ref{fig:Accuracy-(=000025)-increments}. This shows that
the proposed D-CNN model with patch-level pretraining can also learn
effective depth features for object detection, even outperforming
a fine tuned RGB model (3.3\% higher mAP than ZF-net).

\begin{table*}[tbh]
\caption{\label{tab:Comp-RGB-D} Accuracy (\%) on SUN RGB-D}

\centering{}%
\begin{tabular}{ccccccc}
\hline 
\multicolumn{1}{c}{} &
\multirow{2}{*}{Method} &
\multicolumn{2}{c}{CNN models} &
\multicolumn{3}{c}{Accuracy (\%)}\tabularnewline
 &
 & RGB &
Depth &
RGB &
Depth &
RGB-D\tabularnewline
\hline 
\multirow{2}{*}{Baseline} &
Concate. &
\multirow{1}{*}{PL} &
PL &
35.4 &
30.9 &
39.1\tabularnewline
 & Concate. &
\multicolumn{1}{c}{FT-PL} &
FT-PL &
41.5 &
37.5 &
45.4\tabularnewline
 &
Concate. (wSVM) &
FT-PL &
FT-PL &
42.7 &
38.7 &
46.9\tabularnewline
 & Concate.  &
\multirow{1}{*}{R-CNN} &
R-CNN &
44.6 &
41.4 &
47.8\tabularnewline
\hline 
\multirow{5}{*}{\textbf{Proposed}} &
Concate. &
\multirow{1}{*}{FT-PL} &
FT-WSP-ALEX &
- &
37.5 &
48.5\tabularnewline
 & \textbf{RGB-D-CNN} &
FT-PL &
D-CNN &
- &
41.2 &
50.9\tabularnewline
 & \textbf{RGB-D-CNN (wSVM)} &
FT-PL &
D-CNN &
- &
42.4 &
52.4\tabularnewline
 & \textbf{RGB-D-MS (wSVM)} &
FT-PL &
D-CNN &
- &
- &
\textbf{53.4}\tabularnewline
 & \textbf{RGB-D-OB (wSVM)} &
FT-PL+ R-CNN &
D-CNN + R-CNN &
- &
- &
\textbf{53.8}\tabularnewline
\hline 
\multirow{3}{*}{State-of-the-art} &
Zhu et al. \cite{Zhu_2016_CVPR} &
\multicolumn{1}{c}{FT-PL} &
FT-PL &
40.4 &
36.5 &
41.5\tabularnewline
 & Wang et al. \cite{Wang_2016_CVPR} &
FT-PL+ R-CNN &
FT-PL+ R-CNN &
40.4 &
36.5 &
48.1\tabularnewline
 & Song et al. \cite{song2017combining}  &
FT-PL &
FT-PL+ ALEX &
41.5  &
40.1  &
52.3\tabularnewline
\hline 
\multicolumn{7}{c}{FT: Fine tuned, PL: Places-CNN}\tabularnewline
\end{tabular}
\end{table*}

\begin{table}[tbh]
\caption{\label{tab:Comparisons-on-NYUD2}Accuracy (\%) on NYUD2}

\centering{}%
\begin{tabular}{cccc}
\hline 
\multicolumn{1}{c}{} &
\multicolumn{2}{c}{Features} &
\multirow{2}{*}{Acc.}\tabularnewline
Method &
RGB &
Depth & \tabularnewline
\hline 
\multicolumn{4}{c}{Baseline methods}\tabularnewline
\hline 
RGB &
\multirow{1}{*}{FT-PL} &
- &
53.4\tabularnewline
Depth &
\multicolumn{1}{c}{-} &
FT-PL &
51.8\tabularnewline
Concate. &
FT-PL &
FT-PL &
59.5\tabularnewline
RGB &
R-CNN &
- &
52.2\tabularnewline
Depth &
\multirow{1}{*}{-} &
R-CNN &
48.9\tabularnewline
\hline 
\multicolumn{4}{c}{\textbf{Proposed methods}}\tabularnewline
\hline 
\textbf{D-CNN} &
\multicolumn{1}{c}{-} &
D-CNN &
56.4\tabularnewline
\textbf{RGB-D-CNN} &
FT-PL &
D-CNN &
65.1\tabularnewline
\textbf{RGB-D-CNN (wSVM)} &
FT-PL &
D-CNN &
65.8\tabularnewline
\textbf{RGB-D-MS (wSVM)} &
FT-PL &
D-CNN &
\textbf{67.3}\tabularnewline
\textbf{RGB-D-OB (wSVM)} &
FT-PL &
D-CNN &
\textbf{67.5}\tabularnewline
\hline 
\multicolumn{4}{c}{State-of-the-art}\tabularnewline
\hline 
\multicolumn{3}{c}{Gupta \textit{et al.} \cite{Gupta2015} } &
45.4\tabularnewline
\multicolumn{3}{c}{Wang \textit{et al.} \cite{Wang_2016_CVPR} } &
63.9\tabularnewline
\multicolumn{3}{c}{Song \textit{et al.} \cite{song2017combining} } &
66.7\tabularnewline
\hline 
\multicolumn{4}{c}{FT: Fine tuned, PL: Places-CNN}\tabularnewline
\end{tabular}
\end{table}

\subsubsection{RGB-D fusion}

We compare with the performance of RGB-specific, depth-specific and
combined RGB-D models. RGB-D models outperform the modality-specific
ones, as expected (see Table~\ref{tab:Comp-RGB-D}). Places-CNN fine
tuned on RGB still outperforms D-CNN, but just by 0.3\%, showing the
potential of depth features for scene recognition and the proposed
training method. Furthermore, note that the accuracy of D-CNN on depth
data not only outperforms significantly the fine tuned Places-CNN
(by 3.7\%), but this gain is even higher when combined with RGB in
the multimodal case (by 5.5\%). This suggests that depth features
should be learned directly from depth data with a suitable architecture
and training method, and highlights the limitations of transferring
RGB-specific features to depth, even when trained on a large dataset
such as Places. The higher gain also suggests that depth features
learned from scratch are more complementary to RGB ones than those
transferred from Places-CNN.

It is also interesting to compare the gain with respect the strongest
modality-specific network, in this case FT-Places-CNN for RGB (e.g.,
42.7\% with wSVM), and the final multimodal result adding either FT-Places-CNN
(depth) or D-CNN (46.9\% and 52.4\%, respectively). The gain is moderate
in the former (4.2\%), but much higher in the latter (9.7\%), which
further supports that depth features from D-CNN are more complementary
to RGB ones, than those transferred from RGB networks.

Compared with other state-of-the-art methods for RGB-D scene recognition
\cite{Zhu_2016_CVPR,Wang_2016_CVPR}, our method also obtains significantly
higher accuracy by making a more effective use of depth data. These
methods rely on fine tuning Places-CNN on depth data, which we showed
is not desirable because low-level filters remain RGB-specific. Zhu
\textit{et al.} \cite{Zhu_2016_CVPR} learn discriminative RGB-D fusion
layers that help to exploit the redundancy at high layers by aligning
RGB and depth representations. This high-level redundancy largely
results from the fact that both RGB and depth branches derive from
Places-CNN and its RGB-specific features. In contrast, we avoid this
unnecessary redundancy in the first place by learning discriminative
features for depth from the very beginning (i.e., bottom layers),
which are complementary to RGB ones rather than redundant. Wang \textit{et
al.} \cite{Wang_2016_CVPR} extract objects and scene features for
both RGB, depth and surface normals. Despite of exploiting more modalities
and being more expensive computationally (due to object detection),
that framework suffers from the same limitation, and the gain in their
late multimodal fusion phase is largely due that type of redundancy.
Recently, Song \textit{et al.} \cite{song2017combining} achieved
similar performance to ours, by combining three AlexNet networks,
one of them learning directly from depth. However, compared to our
framework, theirs has much more complex models, is inefficient and
has a very complex feature fusion method.

Additionally we also evaluated other RGB-D fusion methods. RGB-D-MS
(wSVM) represents a multi-scale RGB-D fusion variant inspired by \cite{song2017combining},
where we connect lower convolutional layers (such as conv2 and conv3
in Fig.~\ref{fig:RGB-D-scene-recognition}) of D-CNN to the last
fully connected layer with the same operations as in \cite{song2017combining}.
By combining the lower layers of D-CNN, RGB-D-MS outperforms previous
RGB-D-CNN (which only concatenates the last fully connected layers
of RGB and depth CNNs) by 1.0\% in accuracy, and outperforms the state-of-the-art
work \cite{song2017combining} by 1.1\%. The improvement of RGB-D-MS
mainly benefits from the integrating the lower convolutional layers
of D-CNN, which learns depth-specific patterns as shown in Fig.~\ref{fig:Visual_conv1}
(h). RGB-D-OB (wSVM) represents the fusion of RGB-D-CNN features and
features extracted by object detection, which is inspired by \cite{Wang_2016_CVPR}.
In our implementation, RGB-D-OB combines four types of features:
two global (i.e. fine tuned Places CNN with RGB and D-CNN) and two
local extracted with Faster R-CNN \cite{NIPS2015_5638} (i.e. from
RGB and depth images, using the RGB ZF-net architecture \cite{Zeiler2014}
and the proposed D-CNN, respectively). With object based features,
the RGB-D-OB (wSVM) variant outperforms \cite{song2017combining}
by 1.5\%.

\begin{table*}[tbh]
\caption{\label{tab:Comparisons-short-videos} Accuracy (\%) on ISIA RGB-D
(short videos).}

\centering{}%
\begin{tabular}{c|cccccc}
\hline 
\multirow{2}{*}{}  &
\multirow{2}{*}{Method}  &
\multicolumn{1}{c}{Step 1} &
\multicolumn{1}{c}{Step 2} &
\multicolumn{3}{c}{Accuracy (\%)}\tabularnewline
 &
 &
RGB-D Fusion  &
Temporal Embedding  &
RGB  &
Depth  &
RGB-D\tabularnewline
\hline 
\multirow{2}{*}{Baselines}  &
CNN+AVE  &
-  &
AVE  &
42.9  &
34.9  &
-\tabularnewline
 &
CNN+AVE  &
AVE  &
AVE  &
-  &
-  &
48.1\tabularnewline
\hline 
Other  &
CNN (RGB-D)+LSTM  &
Concatenation  &
LSTM  &
-  &
-  &
48.0\tabularnewline
\hline 
\hline 
\multirow{2}{*}{}  &
\multirow{2}{*}{Method}  &
\multicolumn{1}{c}{Step 1} &
\multicolumn{1}{c}{Step 2} &
\multicolumn{3}{c}{Accuracy (\%)}\tabularnewline
 &
 &
Temporal Embedding  &
RGB-D Fusion  &
RGB  &
Depth  &
RGB-D\tabularnewline
\hline 
\multirow{2}{*}{Others}  &
CNN+LSTM  &
LSTM  &
-  &
43.3  &
36.6  &
-\tabularnewline
 &
CNN+LSTM  &
LSTM  &
Concatenation  &
-  &
-  &
48.7\tabularnewline
\hline 
\multirow{2}{*}{\textbf{Proposed}}  &
\textbf{CNN+LSTM (EtE)}  &
LSTM  &
-  &
44.9  &
38.3  &
-\tabularnewline
 &
\textbf{CNN+LSTM (EtE)}  &
LSTM  &
Concatenation  &
-  &
-  &
\textbf{49.9}\tabularnewline
\hline 
\multicolumn{7}{c}{EtE: end to end, AVE: average pooling}\tabularnewline
\end{tabular}
\end{table*}

\begin{table}[tbh]
\setlength{\tabcolsep}{2pt} \global\long\def\arraystretch{1}

\caption{\label{tab:Comparisons-original} Accuracy (\%) on ISIA RGB-D (full
length videos).}

\centering{}%
\begin{tabular}{cccccc}
\hline 
\multirow{2}{*}{RGB-D }  &
\multirow{2}{*}{TE}  &
\multicolumn{3}{c}{Accuracy (\%)}\tabularnewline
 &
 &
 &
RGB  &
Depth  &
RGB-D\tabularnewline
\hline 
CNN+AVE  &
-  &
AVE  &
52.8  &
47.2  &
-\tabularnewline
CNN+AVE  &
AVE  &
AVE  &
-  &
-  &
56.5\tabularnewline
\hline 
\multirow{2}{*}{Method}  &
\multirow{2}{*}{RGB-D }  &
\multirow{2}{*}{TE}  &
\multicolumn{3}{c}{Accuracy (\%)}\tabularnewline
 &
 &
 &
RGB  &
Depth  &
RGB-D\tabularnewline
\hline 
\textbf{CNN+LSTM (EtE)}  &
-  &
LSTM  &
55.6  &
48.1  &
-\tabularnewline
\textbf{CNN+LSTM (EtE)}  &
LSTM  &
Con.  &
-  &
-  &
\textbf{58.3}\tabularnewline
\hline 
\multicolumn{6}{c}{TE: Temporal Embedding, Con.: Concatenation}\tabularnewline
\multicolumn{6}{c}{EtE: End to End, AVE: Average Pooling}\tabularnewline
\multicolumn{6}{c}{}\tabularnewline
\end{tabular}
\end{table}

\subsection{NYUD2}

We also evaluated our approach on NYUD2 and compared to other representative
works (see Table~\ref{tab:Comparisons-on-NYUD2}). Both methods use
more complex frameworks including explicit scene analysis. Gupta \textit{et
al.} \cite{Gupta2015} rely on segmentation and handcrafted features,
while Wang \textit{et al.} on extracting object detection \cite{Wang_2016_CVPR}.
Despite these more structured representations, our approach also outperforms
them. Handcrafted and transferred features are also more competitive
when the training data is more limited. Despite NYUD2 has fewer training
images, our method still can learn better depth and multimodal representations.
Song \textit{et al.} \cite{song2017combining} achieve slightly better
performance than ours with their three AlexNet framework.

We also extend our RGB-D-CNN model with the tricks of multi-scale
fusion and integration with objected based features on NYUD2. The
extended results on NYUD2 are shown in Table~\ref{tab:Comp-RGB-D}.
Compared to \cite{song2017combining}, the proposed RGB-D-MS (wSVM)
obtains a gain of 0.6\%, and RGB-D-OB (wSVM) obtains a gain of 0.8\%.
It is more fair to compare RGB-D-OB (wSVM) with \cite{Wang_2016_CVPR},
since both works integrate global features (extracted from CNN model)
and local features (extracted with object detection). RGB-D-OB (wSVM)
outperforms \cite{Wang_2016_CVPR} by a clear margin of 3.9\%.

\subsection{ISIA RGB-D}

We evaluated the proposed approach on videos, in order to study how
accumulating temporal information helps recognition.

\subsubsection{Data preprocessing}

One problem with the capture of RGB-D videos is that RGB frames are
often blurry, and we found it affected the performance. In order to
alleviate this problem, for our experiments we use a quasilinear sampling
strategy in which we select the least blurred frame every segment
of 5 frames (i.e., sampling approximately 3 frames/s). The degree
of blur is measured as the mean value of the gradient between pixels
(the larger, the less blurry).

Depth videos are stored in 8 bit gray scale. We encode depth frames
to 3-channel images using jet color encoding. We chose this encoding
in this case since it has been shown that the performance is comparable
to HHA encoding while being much faster to compute \cite{eitel2015multimodal}.

\subsubsection{Impact of video length}

The number of frames integrated in the recognition process is an important
parameter. Thus, we first evaluate its impact on the recognition performance
for RGB and depth modalities. We compare two methods to integrate
temporal information: average pooling of the predictions obtained
by the CNN network (AVE) and feeding frame CNN features to a LSTM
network (LSTM). We further include a variant where both CNNs and LSTMs
are trained end-to-end (EtE).

The results are shown in Fig.~\ref{fig:Comparison-temporal-embedding}.
While for RGB the gain is very marginal, for depth the gain is significantly
higher and increasing with the number of frames. Since the range of
RGB cameras is much larger than that of depth ones, the additional
RGB frames do not provide much more additional information, in contrast
to additional depth frames which contain new information that cannot
be captured in each frame separately. This trend is observed in the
three methods evaluated, with LSTM outperforming average pooling,
in particular when using end-to-end training. Particularly, we also
show the detailed comparisons between recognition with images (i.e.,
one frame videos) and videos with 18 frames in Fig.~\ref{fig:Accuracy-(=000025)-increments}.
Each bar in Fig.~\ref{fig:Accuracy-(=000025)-increments} indicate
the gain of using videos, comparing to using images (one frame videos),
for scene recognition with depth data. It can be observed that mAP
(\%) of most object categories are improved when recognizing with
videos with multiple frames. The clear margin of average gain (about
6.3\% in mAP) illustrates the efficiency of recognizing scenes with
depth video data.

\subsubsection{RGB-D recognition}

We evaluated the different variants on both short and full length
videos. For the former, from each video we sample several short segments,
resulting in more training samples but with fewer frames each. We
compared different variants where the order of RGB-D fusions and temporal
integration (i.e., temporal embedding) are different.

The short videos are basically segments of 9 key frames sampled from
the full length videos. The results are shown in Table~\ref{tab:Comparisons-short-videos}.
As in previous datasets, the RGB network has higher performance than
the depth network. Average pooling is outperformed by LSTM based methods,
with end-to-end training helping more. The order of temporal embedding
and fusion is important, and in our experiments the best results are
obtained when the temporal embedding is performed before the RGB-D
fusion. Interestingly, end-to-end fine tuning decreases slightly the
accuracy of RGB only recognition, while increasing significantly that
of depth and RGB-D.

The evaluation on long videos is shown in Table~\ref{tab:Comparisons-original}.
The results are very similar, with the best result obtained by combining
both modalities after modality-specific temporal embedding and end-to-end
training. Accuracies are higher than in the case of short videos
because the network can integrate more temporal information during
both training and inference.

\section{Conclusion}

Compared to RGB, learning genuine depth features is challenging due
to the limited data available and the limited information captured
by the limited range of depth sensors. This problem is usually tackled
using transfer learning, from a deep RGB network trained on a large
dataset (e.g., Places) and then fine tuning with the target depth data. While
effective for the RGB modality, it has significant limitations for
depth data that we highlight in this work. The most important is that
low level filters remain RGB-specific and cannot capture depth-specific
patterns.

We use a radically different approach by focusing mainly on learning
good low level depth-specific filters. A smaller architecture and
a weakly supervised pretraining strategy for the bottom layers enables
us to overcome the problem of very limited depth data. In this way,
the network captures patterns that fine tuned RGB networks are not
able to, and these patterns will be more complementary to RGB ones
in the joint RGB-D network.

Integrating temporal information is particularly helpful for depth
data, capturing in this way information about both near and distant
objects. This is not possible in just one image with current depth sensors.
We studied this case in a new multimodal video scene recognition dataset.

In general, our results show that the proposed training strategy and
spatio-temporal model can exploit much better the depth modality,
with a significantly higher gain over RGB-only scene recognition than
in previous works. We hope this work and dataset can motivate further
research in these directions.

\section*{Acknowledgment}

This work was supported in part by the National Natural Science Foundation
of China under Grant 61532018, in part by the Lenovo Outstanding Young
Scientists Program, in part by National Program for Special Support
of Eminent Professionals and National Program for Support of Top-notch
Young Professionals, in part by the National Postdoctoral Program
for Innovative Talents under Grant BX201700255, and in part by European
Union research and innovation program under the Marie Sk\l odowska-Curie
grant agreement No. 6655919.

\bibliographystyle{IEEEtran}
\bibliography{rgbd}

\begin{thebibliography}{10}
\providecommand{\url}[1]{#1}
\csname url@samestyle\endcsname
\providecommand{\newblock}{\relax}
\providecommand{\bibinfo}[2]{#2}
\providecommand{\BIBentrySTDinterwordspacing}{\spaceskip=0pt\relax}
\providecommand{\BIBentryALTinterwordstretchfactor}{4}
\providecommand{\BIBentryALTinterwordspacing}{\spaceskip=\fontdimen2\font plus
\BIBentryALTinterwordstretchfactor\fontdimen3\font minus
  \fontdimen4\font\relax}
\providecommand{\BIBforeignlanguage}[2]{{%
\expandafter\ifx\csname l@#1\endcsname\relax
\typeout{** WARNING: IEEEtran.bst: No hyphenation pattern has been}%
\typeout{** loaded for the language `#1'. Using the pattern for}%
\typeout{** the default language instead.}%
\else
\language=\csname l@#1\endcsname
\fi
#2}}
\providecommand{\BIBdecl}{\relax}
\BIBdecl

\bibitem{Zhou2014}
B.~Zhou, A.~Lapedriza, J.~Xiao, A.~Torralba, and A.~Oliva, ``Learning deep
  features for scene recognition using places database,'' in \emph{Advances in
  Neural Information Processing Systems ({NIPS})}, 2014, pp. 487--495.

\bibitem{Krizhevsky2012}
A.~Krizhevsky, I.~Sutskever, and G.~E. Hinton, ``Imagenet classification with
  deep convolutional neural networks,'' in \emph{Advances in Neural Information
  Processing Systems ({NIPS})}, 2012, pp. 1106--1114.

\bibitem{Simonyan2015}
K.~Simonyan and A.~Zisserman, ``Very deep convolutional networks for
  large-scale image recognition,'' in \emph{International Conference on
  Learning Representations ({ICLR})}, 2015.

\bibitem{Quattoni2009}
A.~Quattoni and A.~Torralba, ``Recognizing indoor scenes,'' in \emph{{IEEE}
  Conference on Computer Vision and Pattern Recognition ({CVPR})}, 2009.

\bibitem{Xiao2010}
J.~Xiao, J.~Hayes, K.~Ehringer, A.~Olivia, and A.~Torralba, ``{SUN} database:
  Largescale scene recognition from abbey to zoo,'' in \emph{{IEEE} Conference
  on Computer Vision and Pattern Recognition ({CVPR})}, 2010.

\bibitem{Jeff2014}
J.~Donahue, Y.~Jia, O.~Vinyals, J.~Hoffman, N.~Zhang, E.~Tzeng, and T.~Darrell,
  ``{DeCAF}: A deep convolutional activation feature for generic visual
  recognition,'' in \emph{International Conference on Machine Learning
  ({ICML})}, 2014.

\bibitem{Silberman2011}
N.~Silberman and R.~Fergus, ``Indoor scene segmentation using a structured
  light sensor,'' in \emph{International Conference on Computer Vision ({ICCV})
  Workshops}, Nov. 2011, pp. 601--608.

\bibitem{Song2015a}
S.~Song, S.~P. Lichtenberg, and J.~Xiao, ``{SUN RGB-D}: A {RGB-D} scene
  understanding benchmark suite,'' in \emph{{IEEE} Conference on Computer
  Vision and Pattern Recognition ({CVPR})}, Jun. 2015, pp. 567--576.

\bibitem{Wang_2016_CVPR}
A.~Wang, J.~Cai, J.~Lu, and T.-J. Cham, ``Modality and component aware feature
  fusion for {RGB-D} scene classification,'' in \emph{{IEEE} Conference on
  Computer Vision and Pattern Recognition ({CVPR})}, June 2016.

\bibitem{Zhu_2016_CVPR}
H.~Zhu, J.-B. Weibel, and S.~Lu, ``Discriminative multi-modal feature fusion
  for {RGBD} indoor scene recognition,'' in \emph{{IEEE} Conference on Computer
  Vision and Pattern Recognition ({CVPR})}, June 2016.

\bibitem{Gupta_2016_CVPR}
S.~Gupta, J.~Hoffman, and J.~Malik, ``Cross modal distillation for supervision
  transfer,'' in \emph{{IEEE} Conference on Computer Vision and Pattern
  Recognition ({CVPR})}, June 2016.

\bibitem{wang2016correlated}
Z.~Wang, R.~Lin, J.~Lu, J.~Feng \emph{et~al.}, ``Correlated and individual
  multi-modal deep learning for rgb-d object recognition,'' \emph{arXiv
  preprint}, 2016.

\bibitem{eitel2015multimodal}
A.~Eitel, J.~T. Springenberg, L.~Spinello, M.~Riedmiller, and W.~Burgard,
  ``Multimodal deep learning for robust {RGB-D} object recognition,'' in
  \emph{International Conference on Intelligent Robots and Systems ({IROS})},
  2015, pp. 681--687.

\bibitem{Song_2017_AAAI}
X.~Song, L.~Herranz, and S.~Jiang, ``Depth {CNNs} for {RGB-D} scene
  recognition: Learning from scratch better than transferring from rgb-cnns,''
  in \emph{{AAAI} Conference on Artificial Intelligence}, 2017, pp. 4271--4277.

\bibitem{Gupta2015}
S.~Gupta, P.~Arbel{\'a}ez, R.~Girshick, and J.~Malik, ``Indoor scene
  understanding with {RGB-D} images: Bottom-up segmentation, object detection
  and semantic segmentation,'' \emph{International Journal of Computer Vision},
  vol. 112, no.~2, pp. 133--149, 2015.

\bibitem{Banica_2015_CVPR}
D.~Banica and C.~Sminchisescu, ``Second-order constrained parametric proposals
  and sequential search-based structured prediction for semantic segmentation
  in rgb-d images,'' in \emph{{IEEE} Conference on Computer Vision and Pattern
  Recognition ({CVPR})}, June 2015.

\bibitem{Socher2012}
R.~Socher, B.~Huval, B.~Bath, C.~D. Manning, and A.~Y. Ng,
  ``Convolutional-recursive deep learning for {3D} object classification,'' in
  \emph{Advances in Neural Information Processing Systems ({NIPS})}, 2012.

\bibitem{GuptaECCV2014}
S.~Gupta, R.~Girshick, P.~Arbelaez, and J.~Malik, ``Learning rich features from
  {RGB-D} images for object detection and segmentation,'' in \emph{European
  Conference on Computer Vision ({ECCV})}, 2014.

\bibitem{Durand_2016_CVPR}
T.~Durand, N.~Thome, and M.~Cord, ``{WELDON}: Weakly supervised learning of
  deep convolutional neural networks,'' in \emph{{IEEE} Conference on Computer
  Vision and Pattern Recognition ({CVPR})}, June 2016.

\bibitem{Bilen_2016_CVPR}
H.~Bilen and A.~Vedaldi, ``Weakly supervised deep detection networks,'' in
  \emph{{IEEE} Conference on Computer Vision and Pattern Recognition ({CVPR})},
  June 2016.

\bibitem{Oquab_2015_CVPR}
M.~Oquab, L.~Bottou, I.~Laptev, and J.~Sivic, ``Is object localization for
  free? - weakly-supervised learning with convolutional neural networks,'' in
  \emph{{IEEE} Conference on Computer Vision and Pattern Recognition ({CVPR})},
  June 2015.

\bibitem{Rasiwasia2009}
N.~Rasiwasia and N.~Vasconcelos, ``Holistic context modeling using semantic
  co-occurrences,'' in \emph{{IEEE} Conference on Computer Vision and Pattern
  Recognition ({CVPR})}, 2009, pp. 1889--1895.

\bibitem{Rasiwasia2012}
------, ``Holistic context models for visual recognition,'' \emph{{IEEE}
  Transactions on Pattern Analysis and Machine Intelligence}, vol.~34, no.~5,
  pp. 902--917, 2012.

\bibitem{Song2015}
X.~Song, S.~Jiang, and L.~Herranz, ``Joint multi-feature spatial context for
  scene recognition on the semantic manifold,'' in \emph{{IEEE} Conference on
  Computer Vision and Pattern Recognition ({CVPR})}, June 2015.

\bibitem{Song2016}
X.~Song, S.~Jiang, L.~Herranz, Y.~Kong, and K.~Zheng, ``Category co-occurrence
  modeling for large scale scene recognition,'' \emph{Pattern Recognition},
  vol.~59, pp. 98 -- 111, 2016.

\bibitem{wang2017weakly}
Z.~Wang, L.~Wang, Y.~Wang, B.~Zhang, and Y.~Qiao, ``Weakly supervised
  patchnets: Describing and aggregating local patches for scene recognition,''
  \emph{{IEEE} Transactions on Image Processing}, vol.~26, no.~4, pp.
  2028--2041, 2017.

\bibitem{song2017multi}
X.~Song, S.~Jiang, and L.~Herranz, ``Multi-scale multi-feature context modeling
  for scene recognition in the semantic manifold,'' \emph{{IEEE} Transactions
  on Image Processing}, vol.~26, no.~6, pp. 2721--2735, 2017.

\bibitem{Feichtenhofer_2017_CVPR}
C.~Feichtenhofer, A.~Pinz, and R.~P. Wildes, ``Temporal residual networks for
  dynamic scene recognition,'' in \emph{{IEEE} Conference on Computer Vision
  and Pattern Recognition ({CVPR})}, July 2017.

\bibitem{Derpanis_2012_CVPR}
K.~G. Derpanis, M.~Lecce, K.~Daniilidis, and R.~P. Wildes, ``Dynamic scene
  understanding: The role of orientation features in space and time in scene
  classification,'' in \emph{{IEEE} Conference on Computer Vision and Pattern
  Recognition ({CVPR})}, 2012.

\bibitem{Shroff_2010_CVPR}
N.~Shroff, P.~Turaga, and R.~Chellappa, ``Moving vistas: Exploiting motion for
  describing scenes,'' in \emph{{IEEE} Conference on Computer Vision and
  Pattern Recognition ({CVPR})}, June 2010, pp. 1911--1918.

\bibitem{xiao2013sun3d}
J.~Xiao, A.~Owens, and A.~Torralba, ``Sun3d: A database of big spaces
  reconstructed using sfm and object labels,'' in \emph{International
  Conference on Computer Vision ({ICCV})}, 2013, pp. 1625--1632.

\bibitem{Silberman2012}
N.~Silberman, D.~Hoiem, P.~Kohli, and R.~Fergus, ``Indoor segmentation and
  support inference from {RGBD} images,'' in \emph{European Conference on
  Computer Vision ({ECCV})}, ser. ECCV'12, 2012, pp. 746--760.

\bibitem{Karpathy_2014_CVPR}
A.~Karpathy, G.~Toderici, S.~Shetty, T.~Leung, R.~Sukthankar, and L.~Fei-Fei,
  ``Large-scale video classification with convolutional neural networks,'' in
  \emph{{IEEE} Conference on Computer Vision and Pattern Recognition ({CVPR})},
  June 2014, pp. 1725--1732.

\bibitem{Simonyan_2014_NIPS}
K.~Simonyan and A.~Zisserman, ``Two-stream convolutional networks for action
  recognition in videos,'' in \emph{Advances in Neural Information Processing
  Systems ({NIPS})}, Z.~Ghahramani, M.~Welling, C.~Cortes, N.~D. Lawrence, and
  K.~Q. Weinberger, Eds., 2014, pp. 568--576.

\bibitem{Feichtenhofer_2016_CVPR}
C.~Feichtenhofer, A.~Pinz, and A.~Zisserman, ``Convolutional two-stream network
  fusion for video action recognition,'' in \emph{{IEEE} Conference on Computer
  Vision and Pattern Recognition ({CVPR})}, June 2016.

\bibitem{He_2016_CVPR}
K.~He, X.~Zhang, S.~Ren, and J.~Sun, ``Deep residual learning for image
  recognition,'' in \emph{{IEEE} Conference on Computer Vision and Pattern
  Recognition ({CVPR})}, June 2016.

\bibitem{Lazebnik2006}
S.~Lazebnik, C.~Schmid, and J.~Ponce, ``Beyond bags of features: Spatial
  pyramid matching for recognizing natural scene categories,'' in \emph{{IEEE}
  Conference on Computer Vision and Pattern Recognition ({CVPR})}, 2006.

\bibitem{Pulkit_2014}
P.~Agrawal, R.~Girshick, and J.~Malik, ``Analyzing the performance of
  multilayer neural networks for object recognition,'' in \emph{European
  Conference on Computer Vision ({ECCV})}.\hskip 1em plus 0.5em minus
  0.4em\relax Springer, 2014, pp. 329--344.

\bibitem{song2015joint}
X.~Song, S.~Jiang, and L.~Herranz, ``Joint multi-feature spatial context for
  scene recognition on the semantic manifold,'' in \emph{{IEEE} Conference on
  Computer Vision and Pattern Recognition ({CVPR})}, 2015, pp. 1312--1320.

\bibitem{He_2014}
K.~He, X.~Zhang, S.~Ren, and J.~Sun, ``Spatial pyramid pooling in deep
  convolutional networks for visual recognition,'' in \emph{European Conference
  on Computer Vision ({ECCV})}, 2014.

\bibitem{Wang2015}
A.~Wang, J.~Lu, J.~Cai, T.-J. Cham, and G.~Wang, ``Large-margin multi-modal
  deep learning for rgb-d object recognition,'' \emph{{IEEE} Transactions on
  Multimedia}, vol.~17, 2015.

\bibitem{Zeiler2014}
M.~D. Zeiler and R.~Fergus, ``Visualizing and understanding convolutional
  networks,'' in \emph{European Conference on Computer Vision ({ECCV})}, Cham,
  2014, pp. 818--833.

\bibitem{NIPS2015_5638}
S.~Ren, K.~He, R.~Girshick, and J.~Sun, ``Faster {R-CNN}: Towards real-time
  object detection with region proposal networks,'' in \emph{Advances in Neural
  Information Processing Systems ({NIPS})}, C.~Cortes, N.~D. Lawrence, D.~D.
  Lee, M.~Sugiyama, and R.~Garnett, Eds., 2015, pp. 91--99.

\bibitem{song2017combining}
X.~Song, S.~Jiang, and L.~Herranz, ``Combining models from multiple sources for
  {RGB-D} scene recognition,'' in \emph{International Joint Conference on
  Artificial Intelligence ({IJCAI})}, 2017, pp. 4523--4529.

\end{thebibliography}

\end{document}